\PassOptionsToPackage{unicode}{hyperref}
\PassOptionsToPackage{hyphens}{url}
\documentclass[
  11pt,
]{article}
\usepackage{xcolor}
\usepackage[margin=1in]{geometry}
\usepackage{amsmath,amssymb}
\usepackage[T1]{fontenc}
\usepackage[utf8]{inputenc}
\usepackage{textcomp}
\usepackage{tgtermes}

\IfFileExists{upquote.sty}{\usepackage{upquote}}{}
\IfFileExists{microtype.sty}{
  \usepackage[]{microtype}
  \UseMicrotypeSet[protrusion]{basicmath} 
}{}
\makeatletter
\@ifundefined{KOMAClassName}{
  \IfFileExists{parskip.sty}{%
    \usepackage{parskip}
  }{
    \setlength{\parindent}{0pt}
    \setlength{\parskip}{6pt plus 2pt minus 1pt}}
}{
  \KOMAoptions{parskip=half}}
\makeatother
\usepackage{longtable,booktabs,array}
\usepackage{calc} 
\usepackage{etoolbox}
\makeatletter
\patchcmd\longtable{\par}{\if@noskipsec\mbox{}\fi\par}{}{}
\makeatother
\IfFileExists{footnotehyper.sty}{\usepackage{footnotehyper}}{\usepackage{footnote}}
\makesavenoteenv{longtable}
\usepackage{graphicx}
\makeatletter
\newsavebox\pandoc@box
\newcommand*\pandocbounded[1]{
  \sbox\pandoc@box{#1}%
  \Gscale@div\@tempa{\textheight}{\dimexpr\ht\pandoc@box+\dp\pandoc@box\relax}%
  \Gscale@div\@tempb{\linewidth}{\wd\pandoc@box}%
  \ifdim\@tempb\p@<\@tempa\p@\let\@tempa\@tempb\fi
  \ifdim\@tempa\p@<\p@\scalebox{\@tempa}{\usebox\pandoc@box}%
  \else\usebox{\pandoc@box}%
  \fi%
}
\def\fps@figure{htbp}
\makeatother
\NewDocumentCommand\citeproctext{}{}
\NewDocumentCommand\citeproc{mm}{%
  \begingroup\def\citeproctext{#2}\cite{#1}\endgroup}
\makeatletter
 \let\@cite@ofmt\@firstofone
 \def\@biblabel#1{}
 \def\@cite#1#2{{#1\if@tempswa , #2\fi}}
\makeatother
\newlength{\cslhangindent}
\newlength{\csllabelwidth}
\newenvironment{CSLReferences}[2] 
 {\begin{list}{}{%
  \setlength{\itemindent}{0pt}
  \setlength{\leftmargin}{0pt}
  \setlength{\parsep}{0pt}
  \ifodd #1
   \setlength{\leftmargin}{\cslhangindent}
   \setlength{\itemindent}{-1\cslhangindent}
  \fi
  \setlength{\itemsep}{#2\baselineskip}}}
 {\end{list}}
\usepackage{calc}

\providecommand{\tightlist}{%
  \setlength{\itemsep}{0pt}\setlength{\parskip}{0pt}}

\usepackage{booktabs}
\usepackage{float}
\usepackage{etoolbox}
\usepackage{amsmath,amssymb}
\usepackage{microtype}
\usepackage[font=small,labelfont=bf]{caption}
\usepackage{setspace}
\usepackage[title,titletoc]{appendix}
\usepackage{tikz}
\usetikzlibrary{arrows.meta,positioning,calc}
\makeatletter\renewcommand\@fnsymbol[1]{}\makeatother
\renewcommand{\topfraction}{0.9}\renewcommand{\bottomfraction}{0.7}\renewcommand{\textfraction}{0.05}\renewcommand{\floatpagefraction}{0.85}
\usepackage{bookmark}
\IfFileExists{xurl.sty}{\usepackage{xurl}}{} 
\hypersetup{
  pdftitle={Propose{,} Don't Judge: An Anytime-Valid Referee for LLM Agents That Mine Investment Factors},
  pdfauthor={Bo Qu\^{}\{1,*\}, Mingguang Chen\^{}\{1\}, Licheng Wang\^{}\{2\}},
  hidelinks,
  pdfcreator={LaTeX via pandoc}}

\title{Propose, Don't Judge: An Anytime-Valid Referee for LLM Agents
That Mine Investment Factors}
\author{Bo Qu\(^{1,*}\), Mingguang Chen\(^{1}\), Licheng
Wang\(^{2}\)\thanks{$^{*}$Corresponding author. $^{1}$DeepGrounding, deepgrounding.org, deepgroundingai@gmail.com. $^{2}$AlphaAvatar, alphaavatar.ai.}}
\date{}

\begin{document}
\maketitle
\begin{abstract}
Language-model agents now run the whole of quantitative factor research:
they propose investment factors, backtest them, select the survivors and
retire them. We ask which of those jobs an agent should keep. Our answer
is \emph{governed self-evolution}: the agent may propose, and a frozen
statistical referee that the agent cannot touch must judge. The referee
scores each candidate only on market outcomes revealed after submission,
by betting, so its false-discovery guarantee holds at every stopping
time for any proposal policy. We cross three proposers (a script, a
bandit and a language model) with this referee and with three
deliberately leaky ones, in a synthetic world with planted truth, a
probe-authoring environment and a ten-year walk-forward on the CSI 500.
Who judges sets the number of false admissions: the frozen referee
admits 5--11\(\times\) fewer sub-threshold factors than the leaky
referees under a scripted proposer, and no proposer closes that gap. Who
proposes sets the yield: the language model beats the script, matches
the bandit, and adds the one capability a bandit lacks, writing its own
diagnostic probes. The certificate's price is time: an admitted true
factor waits about 500 trading days, and the certified portfolio's
Sharpe ratio therefore trails an ungated one. Judging belongs to the
procedure; proposing and instrument-making belong to the agent.
\end{abstract}

\AtBeginEnvironment{tabular}{\small}

\section{Introduction}\label{sec:intro}

Language-model agents are taking over quantitative factor research. By
our count, about twenty-five agentic factor-mining systems appeared in
the eighteen months to mid-2026. In every system we examined, the agent
performs every job in the pipeline. It generates candidate factors,
backtests them, selects the survivors, and decides when to drop them.
This paper asks which jobs the agent should keep. We answer by measuring
a division of labour. The agent may propose. A statistical procedure
that the agent cannot touch must judge.

This division matters because evidence about a factor arrives slowly. An
\emph{investment factor} scores every stock on a characteristic and bets
that the ranking predicts returns. The bet buys the top fifth, sells the
bottom fifth, holds, and repeats. Its daily report card is the
\emph{rank-IC}. This is the rank correlation between yesterday's ranking
and today's returns across the universe. In a 500-stock universe, a good
factor averages two or three hundredths a day. Its standard deviation is
several times larger. A factor therefore earns trust only over years. A
search over hundreds of candidates finds many that look good by luck.

An agent in a loop creates two errors that today's evaluations were not
designed to control. The systems we examined apply a fixed checklist to
the agent's output: a \(t\)-statistic, an information ratio, or, at
best, a deflated Sharpe ratio. Those checklists assume a batch of tests
fixed in advance. An agent is not a fixed batch if it reads each verdict
before choosing its next proposal and never stops. It resubmits until
something passes. This is \emph{optional stopping}. It passes
near-duplicates of a lucky candidate together. This is
\emph{multiplicity}. A correction computed in advance requires a
proposal process known in advance. A learning agent has no such process.

No prior system separates proposing from judging. The one line of prior
work that puts an anytime-valid test around an agent certifies the
agent's own modifications, not the hypotheses it proposes
(\citeproc{ref-sea2026}{Sengupta 2026}; \citeproc{ref-pace2026}{Shawn
2026}). A survey of self-improving systems reaches the same gap from the
other side: every improvement loop rests on a verification signal, and
governance-grade measurement of that loop is underdeveloped
(\citeproc{ref-chen2026rsi}{Chen et al. 2026}). The open question is
therefore \textbf{which jobs in a factor lifecycle benefit from an
agent's intelligence, and which must be governed by a statistical
procedure the agent cannot touch}. We call the resulting arrangement
\emph{governed self-evolution}. The agent's surface may evolve without
limit: its proposals, its diagnoses, and the instruments it writes for
itself. A frozen statistical referee governs what enters the portfolio.

The referee scores a candidate only on market outcomes revealed after
submission. We split the pipeline into a \emph{proposer} and a
\emph{referee}, then freeze the referee. Each candidate factor is scored
on the rank-IC from the days after submission. Nothing the proposer saw
before submission enters the score. The score is a bet. Each day, the
referee stakes part of the candidate's capital on that day's rank-IC
being positive. A factor with no edge rarely produces capital that grows
large. A process with this property is called an \emph{e-process}.

Two further rules act on that capital. The admission rule asks whether a
candidate has any edge. A candidate is \emph{admitted} when its capital
clears a bar accounting for every candidate ever submitted. That rule is
\emph{online e-BH}, the e-value form of the Benjamini--Hochberg
procedure. The retirement rule asks whether a deployed factor's edge
still covers its own trading costs. The registered viability threshold
\(\delta\) is the edge that just pays. A factor is \emph{retired} when a
second bet against it shows that its edge has fallen below \(\delta\).
That rule is an \emph{e-detector}, the e-value form of a changepoint
chart. Section \ref{sec:referee} builds all three procedures, and
Appendix \ref{app:glossary} collects their definitions.

The guarantee holds regardless of who proposes. Every bet is sized
before the day's outcome is seen. The market reveals outcomes regardless
of the proposer's actions. The false-discovery rate over admissions is
therefore controlled at every stopping time. The retirement detector's
run length to a false alarm is bounded below. Both statements hold for
any proposer, including one that rewrites its own agent surface
(\S\ref{sec:validity}). They rely on one hypothesis: the proposer has no
information about the post-submission stream. A live run satisfies this
hypothesis. A historical replay by a model trained on the period may
not.

The proposer can therefore be anything, and we test a ladder of three. A
\emph{round-robin script} cycles through the factor families in a fixed
order. A \emph{discounted-UCB bandit} is the standard multi-armed-bandit
allocator. It sends more submissions to families with recent successes
and slowly forgets. A \emph{language-model controller} also allocates
submissions. It diagnoses retirements and writes its own
\emph{perception probes}. These small diagnostic programs measure a
suspected fault and never enter a statistical test.

The experiment separately measures the effects of who judges and who
proposes. Besides the valid referee, we run three deliberately leaky
referees that each mimic a common practice. Crossing the three proposers
with the four referees measures each effect while holding the other
fixed. Three environments share this design. A synthetic world with
planted truth isolates the mechanism. A probe-authoring environment with
held-out fault types measures the one agent capability we claim. A
ten-year point-in-time walk-forward on the CSI 500 tests which results
survive real data. It also follows the certified factors into a
portfolio after costs. Every comparison uses the same random draws, and
model families are never pooled.

\textbf{What we find.} A \emph{campaign} is one walk-forward run of the
pipeline, lasting seven to ten years depending on the start year.
Real-data counts are per campaign. A \emph{false admission} is an
admitted factor whose realised edge falls short of the viability
threshold.

\emph{First, who judges sets the number of false admissions, and no
proposer substitutes for it.} In the synthetic world, the frozen referee
admits no false factor in any seed. The three leaky referees admit
\(0.26\)--\(0.85\) per submission. The figures are the same for script,
bandit, and language model (Table \ref{tbl:synthetic}). On real data,
the frozen referee admits 11.7 false factors per campaign. The leaky
referees admit 86--196 under the script and 37--78 under the language
model. The frozen referee's advantage is 5--11\(\times\) under the
script and 2.6--4.9\(\times\) under the language model, based only on
days after admission. A patient desk that waits 500 days and then
corrects for multiplicity comes close where families are clearly good or
bad. It is four times less accurate where many edges are small
(\S\ref{sec:comparator}).

\emph{Second, who proposes sets the yield, and the language model's
value is real and bounded.} The language model finds more true factors
than the script in six of six family \(\times\) library settings, in
every start year. It is level with the bandit in most settings. It falls
behind in one family on the nine-family library. Its one capability a
bandit lacks is instrument-making. Authored probes significantly lower
intervention regret in three of six evaluable model families. In the
other three, they are indistinguishable from a fixed diagnosis menu, so
authoring never hurts (Table \ref{tbl:probe}).

\emph{Third, the certificate has a price, paid in waiting and in what
the bar selects.} The certified portfolio does not lose money at the
registered cost. Its alpha cannot be told from zero. Its Sharpe ratio
net of costs is below the ungated portfolio's. Two mechanisms account
for the gap. An admitted true factor waits about 500 trading days. The
ungated pipeline is invested during that wait. The strongest daily
statistic clears the bar fastest. In this library, that statistic is
short-horizon reversal, the family that does not pay after costs. A
fixed-horizon test with the same patience therefore builds a better
portfolio (\S\ref{sec:portfolio}).

\textbf{Contributions.} We propose and measure governed self-evolution,
a framework pairing an evolving agent surface with a frozen statistical
referee.

\begin{itemize}
\tightlist
\item
  (C1) \emph{A referee that can be frozen.} It scores each candidate
  only on data revealed after submission. The false-discovery rate over
  admitted factors therefore holds at every stopping time, whatever the
  proposer does. We state the conditions under which this holds inside
  an agentic loop.
\item
  (C2) \emph{An experiment that separates the two jobs.} Three proposers
  each run against the valid referee and three leaky ones. Common random
  draws let the two effects be read separately.
\item
  (C3) \emph{A measurement of what the language model adds.} It beats
  the script on yield. It does not reliably beat the bandit on
  allocation. It writes its own diagnostic probes for faults it has
  never seen.
\item
  (C4) \emph{A ten-year real-data study on the CSI 500} that follows the
  certified factors into a portfolio with costs. It accounts for the gap
  between the certified and ungated portfolios.
\end{itemize}

We do not offer an investment system. The portfolio layer is a measuring
instrument. The paper has three stages: the pipeline
(\S\ref{sec:referee}--\S\ref{sec:controllers}), the experiment
(\S\ref{sec:design}) and what it found
(\S\ref{sec:results}--\S\ref{sec:discussion}).

\section{Related work}\label{sec:related}

Agentic factor mining treats the agent as the whole pipeline. Research
spans reinforcement-learning generators over an operator grammar
(\citeproc{ref-alphagen2023}{Yu et al. 2023}), agentic loops with
critique and refinement (\citeproc{ref-rdagent2025}{Li et al. 2025};
\citeproc{ref-alphaagent2025}{Tang et al. 2025}), and
structured-semantics search (\citeproc{ref-alphaschema2026}{Yi et al.
2026}). In each approach, the same agent generates, evaluates, and
selects. Evaluation uses held-out IC, ICIR, or Sharpe after a fixed
mining budget. At most, a static deflation is applied to a batch
(\citeproc{ref-bailey2014dsr}{Bailey and López de Prado 2014}). None
controls the sequential error created by the loop. The closest evidence
is a boundary finding. AlphaSchema reports that model identity has
little effect within a structured search
(\citeproc{ref-alphaschema2026}{Yi et al. 2026}). Our proposer ladder
reproduces that finding.

Anytime-valid inference supplies every piece the referee needs. We apply
it without extending it. Test supermartingales and Ville's inequality
(\citeproc{ref-ville1939}{Ville 1939}) permit a test that can be read
every day. We use betting e-processes for bounded means
(\citeproc{ref-waudby2024betting}{Waudby-Smith and Ramdas 2024}). The
e-BH procedure (\citeproc{ref-wang2022ebh}{Wang and Ramdas 2022}) and
its online form (\citeproc{ref-fischer2024online}{Fischer et al. 2024};
\citeproc{ref-wang2025stopping}{Wang et al. 2025};
\citeproc{ref-deheide2026}{Heide 2026}) control the false-discovery rate
over a growing set at data-dependent stopping times. E-values merge
under arbitrary dependence (\citeproc{ref-vovk2021merging}{Vovk and Wang
2021}). Compound and weighted e-values (\citeproc{ref-xu2026compound}{Xu
et al. 2026}; \citeproc{ref-sun2026weighted}{Sun and Wang 2026};
\citeproc{ref-clerico2026}{Clerico 2026}) fix the form of the bet. Our
contribution states when this theory survives an agentic loop. It also
measures the cost.

Certified self-evolution is the nearest architecture, but it certifies a
different object. SEA (\citeproc{ref-sea2026}{Sengupta 2026}) and PACE
(\citeproc{ref-pace2026}{Shawn 2026}) gate a self-modifying agent with
an anytime-valid test. Both certify \emph{the agent's own
modifications}. A factor pipeline is harder. The agent proposes external
hypotheses whose truth a market decides. The admitted set grows without
bound. The guarantee must control the false-discovery rate over that
set. The division of labour, the pipeline implementing it, and the
experiment measuring it are ours.

Retirement and controller comparison each draw on established results.
Factor edges decay after discovery (\citeproc{ref-mclean2016}{McLean and
Pontiff 2016}). Many published factors are false from the outset
(\citeproc{ref-harvey2016}{Harvey et al. 2016}). A deployed factor must
therefore be monitored. E-detectors
(\citeproc{ref-shin2024edetectors}{Shin et al. 2024}) provide our
retirement rule. Error-over-patience
(\citeproc{ref-dandapanthula2025eop}{Dandapanthula and Ramdas 2025}) is
the library-level metric. Static replay of logged agent trajectories is
invalid once policies diverge (\citeproc{ref-replaygap2026}{Gonuguntla
2026}). Controllers are therefore compared through branched live
rollouts with common random numbers. Portfolio results are reported
after costs. They also use the deflated Sharpe ratio
(\citeproc{ref-bailey2014dsr}{Bailey and López de Prado 2014}). Table
\ref{tbl:systems} summarises what each line certifies.

\begingroup\small

\begin{table}[htbp]
\centering
\caption{What each line of work certifies and when its guarantee holds.
``Adaptive proposals'' are candidates a learning agent chooses using
earlier candidates' outcomes. ``Growing set'' means controlling the
false-discovery rate over every factor ever admitted. The agentic miners
are (\citeproc{ref-alphagen2023}{Yu et al. 2023};
\citeproc{ref-rdagent2025}{Li et al. 2025};
\citeproc{ref-alphaagent2025}{Tang et al. 2025};
\citeproc{ref-alphaschema2026}{Yi et al. 2026}). Certified
self-modification is (\citeproc{ref-sea2026}{Sengupta 2026};
\citeproc{ref-pace2026}{Shawn 2026}). Notes: \(^a\) denotes a fixed
batch. \(^b\) denotes a bounded modification space. \(^c\) denotes a
per-modification budget. \(^d\) denotes online e-BH. For \(^e\), the
e-detector guarantees the average run length (\S\ref{sec:retirement}).
\(^f\) applies to a proposer with no information about post-submission
outcomes (\S\ref{sec:validity}).}\label{tbl:systems}
\begin{tabular}{@{}
  >{\raggedright\arraybackslash}p{(\linewidth - 10\tabcolsep) * \real{0.2874}}
  >{\raggedright\arraybackslash}p{(\linewidth - 10\tabcolsep) * \real{0.2414}}
  >{\centering\arraybackslash}p{(\linewidth - 10\tabcolsep) * \real{0.1149}}
  >{\centering\arraybackslash}p{(\linewidth - 10\tabcolsep) * \real{0.1149}}
  >{\centering\arraybackslash}p{(\linewidth - 10\tabcolsep) * \real{0.1149}}
  >{\centering\arraybackslash}p{(\linewidth - 10\tabcolsep) * \real{0.1264}}@{}}
\toprule\noalign{}
\begin{minipage}[b]{\linewidth}\raggedright
Approach
\end{minipage} & \begin{minipage}[b]{\linewidth}\raggedright
Certifies
\end{minipage} & \begin{minipage}[b]{\linewidth}\centering
Anytime Valid
\end{minipage} & \begin{minipage}[b]{\linewidth}\centering
Adaptive Proposals
\end{minipage} & \begin{minipage}[b]{\linewidth}\centering
Growing Set
\end{minipage} & \begin{minipage}[b]{\linewidth}\centering
Retirement
\end{minipage} \\
\midrule\noalign{}
Backtest Checklist & A batch of backtests & No & No & No & No \\
Batch Deflation & A batch of trials & No & Partial\(^a\) & No & No \\
Agentic Miners & As the two rows above & No & No & No & Heuristic \\
Certified Self-Modification & The agent's own modifications & Yes &
Partial\(^b\) & No\(^c\) & No \\
This Paper & External hypotheses, scored after submission & Yes &
Yes\(^f\) & Yes\(^d\) & Yes\(^e\) \\
\bottomrule
\end{tabular}
\end{table}

\endgroup

\section{Problem: an adaptive, feedback-coupled factor
stream}\label{sec:problem}

This section defines the objects used throughout the paper. They are the
candidate stream, market outcomes, candidate states, guarantees, outcome
measures, and threats.

We study a stream of candidate factors whose next member depends on
everything observed so far. A library \(\mathbb{L}\) is a finite set of
factor families \(f\in\mathcal{F}\) and their parameter grids. Each
family represents an economic hypothesis, such as momentum or value. A
proposer, called the controller, emits candidate factors
\(\varphi_1,\varphi_2,\dots\). Candidate \(i\) is a pair
\((f_i,\theta_i)\) submitted on trading day \(s_i\). Candidates within a
family are near-duplicates. An adaptive controller resubmits variants of
candidates that passed. The stream is therefore neither independent nor
fixed in advance.

The market reveals outcomes one day at a time. Let \(\mathcal{U}_t\) be
the point-in-time universe on day \(t\). It contains stocks whose index
membership was known at \(t-1\) and that are tradable and not
limit-locked. For day \(t>s_i\), candidate \(i\)'s outcome is
\(X^{(i)}_t\). It is the cross-sectional Spearman rank correlation
between its signal at the close of \(t-1\) and the returns on \(t\),
over \(\mathcal{U}_t\). Outcomes are exogenous. No action on day \(t\)
changes \(X^{(i)}_{t}\).

Anything used on a day must have been known the day before. This rule
has a formal name. Write \(\mathcal{G}_{t-1}\) for the \emph{global
filtration}. It contains everything known at the close of day \(t-1\),
including all market data, candidates' outcomes, controller outputs, and
referee states. Every quantity used on day \(t\) must be measurable with
respect to \(\mathcal{G}_{t-1}\). We call this rule the alignment
invariant (\S\ref{sec:validity}).

\begin{figure}[t]
\centering
\scalebox{0.75}{\begin{tikzpicture}[font=\footnotesize,
  box/.style={draw, rounded corners=3pt, align=center, inner sep=4pt},
  gov/.style={box, fill=blue!10},
  agt/.style={box, fill=orange!18},
  dat/.style={box, fill=gray!12},
  arr/.style={-{Stealth[length=2.2mm]}, semithick},
  fb/.style={-{Stealth[length=2.2mm]}, semithick, dashed, gray!45!black},
  node distance=5mm and 11mm]
\node[dat, text width=74mm] (lib) {\textbf{Library} (Frozen Within an Epoch)\\ Families $\times$ Parameter Grids};
\node[agt, below=of lib, text width=74mm] (ctl) {\textbf{Controller} (Agent Surface, Swappable)\\ Direction: allocate submissions across families\\ Construction: draw a candidate from the grid\\ Diagnosis: label an alarm, or escalate\\ Probe authoring: sandboxed instruments; memory};
\node[gov, below=of ctl, text width=74mm] (ref) {\textbf{Referee} (Trust Kernel, Frozen)\\ Post-Submission e-Process per Candidate\\ $\to$ Online e-BH Admission\\ $\to$ e-Detector Retirement};
\node[dat, below=of ref, text width=74mm] (exe) {\textbf{Execution Layer}\\ Decay Curve $\to$ Holding Frequency, Trade or Shelve\\ Book: turnover cap, index hedge, 15 bp per side};
\node[gov, right=of ref, text width=40mm] (pit) {\textbf{Point-in-Time Data}\\ Daily rank-IC of each candidate, revealed after submission};
\node[gov, below=of pit, text width=40mm] (log) {\textbf{Decision Log} (Append-Only)\\ + Cost Meter; every LLM call recorded};
\draw[arr] (lib) -- node[right]{\tiny Grammar} (ctl);
\draw[arr] (ctl) -- node[right]{\tiny Submit} (ref);
\draw[arr] (ref) -- node[right]{\tiny Admitted Set, Retirements} (exe);
\draw[arr] (pit) -- node[above=1pt]{\tiny Outcomes} (ref);
\draw[fb] (ref.west) to[out=180,in=180] node[left, pos=0.5, align=right]{\tiny Verdicts,\\ \tiny Diagnostics} (ctl.west);
\draw[arr] (ref.east) ++(0,-3mm) -- ++(4mm,0) |- (log.west);
\node[below=2mm of exe, text width=74mm, align=center, font=\scriptsize] (note) {The book never feeds back into the controller\\ (Proposition 2: any book is re-computed from the log)};
\end{tikzpicture}}
\caption{The division of labour. The controller (orange) proposes, diagnoses and writes its own instruments. The referee, the data layer and the decision log (blue) are frozen components the controller cannot mount, patch or intercept. The execution layer (grey) turns admitted factors into positions and is never seen by the referee or the controller.}
\label{fig:arch}
\end{figure}
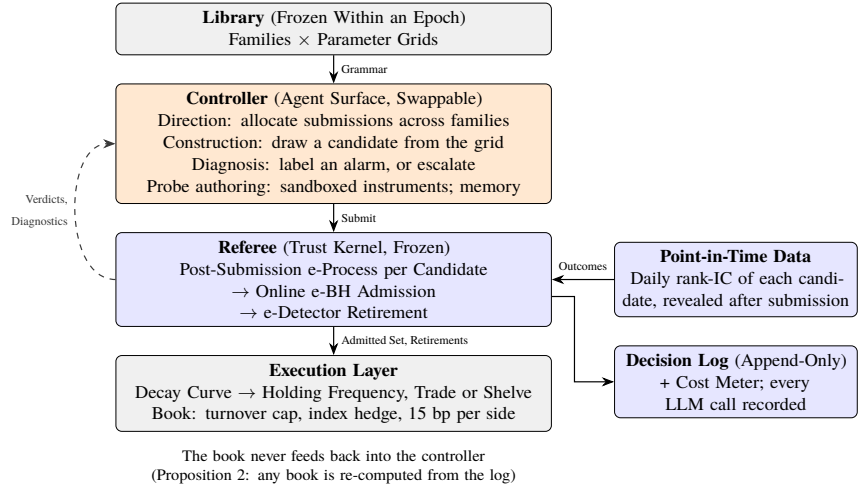

A candidate passes through three states, shown in Figure \ref{fig:arch}.
Only the referee moves it between them. A candidate is \emph{open} until
it is admitted or expires. An admitted candidate is \emph{deployed}. The
detector makes a deployed factor \emph{retired}. The portfolio, called
the \emph{book}, holds every deployed factor.

The referee provides two guarantees. Both must hold at any horizon and
any stopping time. The false-discovery rate over admissions is at most
\(\alpha\). The expected run length to a false retirement is at least
\(A\). Appendix \ref{app:glossary} defines a false retirement and its
clock.

Two outcome measures record what a campaign produced. To compute
\emph{Realised yield per submission}, count factors still deployed at
campaign end whose post-submission mean rank-IC is at least \(\delta\).
Divide this count by the number of candidates submitted. \emph{Realised
false admissions} counts factors ever admitted whose post-submission
mean rank-IC falls short of \(\delta\). Real data have no planted truth.
We therefore label a candidate true when its mean rank-IC from
submission to campaign end is at least \(\delta\). The candidate and its
submission day fix the label. The label is blind to the referee's
decision.

Two properties of that label determine how to read the real-data counts.
First, the label's window starts at submission. It therefore includes
the days on which the referee decided. This overlap favours a referee
that admits late. To remove it, \S\ref{sec:real-referee} re-judges every
admitted factor using only the days after its admission. Second, a
realised false admission is a count against the viability threshold
\(\delta\). The guarantee bounds a false-discovery \emph{proportion}
against a \emph{zero} edge (\S\ref{sec:betting}). An admitted factor
with a small positive edge is therefore a realised false admission, not
an error of the guarantee.

The threat model contains four entries. The referee addresses three by
construction. Optional stopping and resubmitting near-copies are the two
errors of \S\ref{sec:intro}. \emph{Verifier-in-the-loop adaptation}
means that the controller learns what the referee rewards. The fourth
threat is \emph{knowledge-cutoff leakage}. A live run removes it because
no proposer can know the post-submission stream. A historical replay may
not remove it if the model's training covers the campaign years. The
guarantee's hypothesis excludes that case (\S\ref{sec:validity}). The
language-model arms' real-data results are therefore read as replay
outcomes. Appendix \ref{app:glossary} maps each threat to its answer.

\section{The referee}\label{sec:referee}

The referee answers three questions about every candidate factor. Does
this candidate have any edge at all? Which candidates can be admitted
while keeping false admissions low among all candidates ever submitted?
Once deployed, has a factor's edge fallen below the level that pays for
its trading?

A classical test cannot answer them here. It fixes its sample size in
advance and is read once. The referee is read every day over a candidate
set that grows as long as the system runs.

One idea answers all three questions: treat evidence as capital in a
betting game. The game is fair or unfavourable to the candidate whenever
the candidate is in fact null. Capital in such a game cannot be expected
to grow. Therefore, the chance that it ever reaches a high level is
small, whenever one looks. The first question becomes a bet that one
candidate's edge is positive. The second becomes a rule for how high
each candidate's capital must climb when many candidates are betting.
The third becomes a bet against a deployed factor. Its capital grows
only when the edge is gone.

\begin{figure}
\centering
\includegraphics[width=0.92\linewidth,height=\textheight,keepaspectratio,alt={One candidate's life under the referee, on a simulated stream (an illustration of the mechanics, not a reported result). Top: the 63-day mean of the daily rank-IC of a factor whose edge is 0.03 and then decays to zero, and of a null factor, with the viability threshold \textbackslash delta. Middle: the capital of the admission bet on a log scale. The true factor's capital reaches the single-candidate bar 1/\textbackslash alpha drawn here (the bar as run, over N\_v slots, is higher, ) and the factor is admitted; the null factor's capital never does. Bottom: the retirement detector, started at admission, stays flat while the edge is intact and climbs to A once the edge has decayed.}]{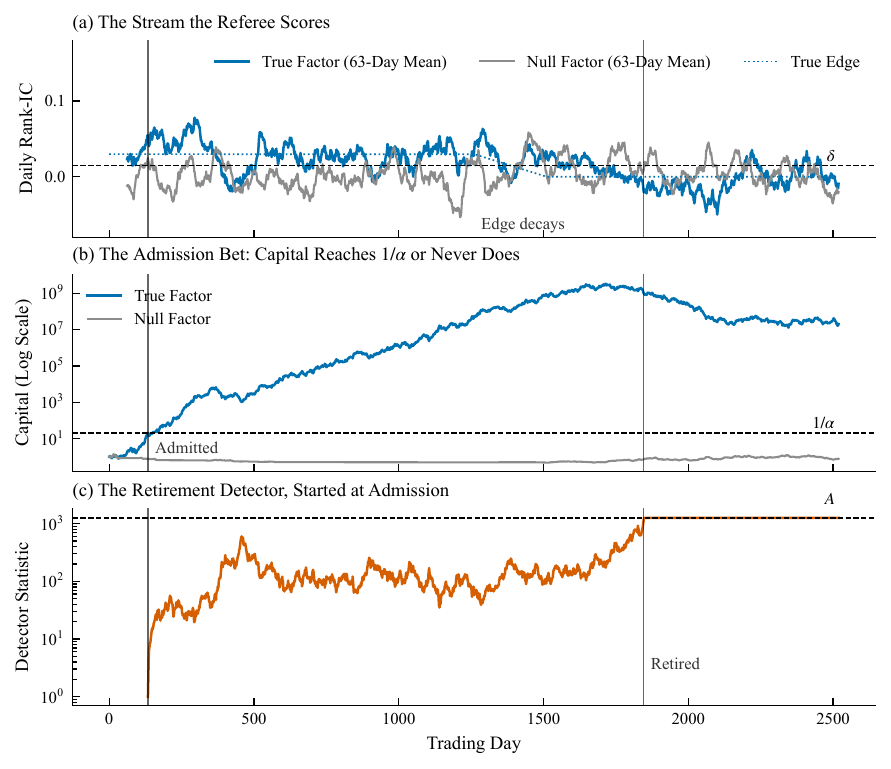}
\caption{One candidate's life under the referee, on a simulated stream
(an illustration of the mechanics, not a reported result). Top: the
63-day mean of the daily rank-IC of a factor whose edge is \(0.03\) and
then decays to zero, and of a null factor, with the viability threshold
\(\delta\). Middle: the capital of the admission bet on a log scale. The
true factor's capital reaches the single-candidate bar \(1/\alpha\)
drawn here (the bar as run, over \(N_v\) slots, is higher,
\S\ref{sec:admission}) and the factor is admitted; the null factor's
capital never does. Bottom: the retirement detector, started at
admission, stays flat while the edge is intact and climbs to \(A\) once
the edge has decayed.}\label{fig:life}
\end{figure}

The section follows that order. Section \ref{sec:toolbox} states the one
property of betting capital underlying all three procedures. Sections
\ref{sec:betting}, \ref{sec:admission} and \ref{sec:retirement} build
those procedures. Figure \ref{fig:life} shows one candidate's life under
them. Section \ref{sec:validity} gives the condition under which an
agent in the loop leaves all three intact.

\subsection{Evidence as capital}\label{sec:toolbox}

A test supermartingale measures evidence and stays honest when read
every day. Write \((\mathcal{G}_t)\) for the global filtration of
\S\ref{sec:problem}, containing everything known at the close of day
\(t\). A nonnegative process \(W\) with \(W_0=1\) is a \emph{test
supermartingale} for a null hypothesis \(H_0\) if
\(\mathbb{E}_{H_0}[W_t\mid\mathcal{G}_{t-1}]\le W_{t-1}\). Under the
null, tomorrow's expected capital is at most today's.

Two consequences support everything that follows. The first is
\emph{optional stopping}: \(\mathbb{E}_{H_0}[W_{\tau}]\le1\) for any
stopping time \(\tau\). Such a stopping time is any rule that uses only
the past to decide when to stop. Capital read at a data-dependent moment
remains a fair quantity, called an \emph{e-value}. The second is
\emph{Ville's inequality} (\citeproc{ref-ville1939}{Ville 1939}):
\(\mathbb{P}_{H_0}(\exists t:\ W_t\ge1/\alpha)\le\alpha\). A null
candidate's capital reaches \(1/\alpha\) with probability at most
\(\alpha\).

\subsection{One candidate: a bet on the post-submission
stream}\label{sec:betting}

The first procedure tests whether one candidate has an edge. It plays
the game of \S\ref{sec:toolbox} on that candidate's post-submission
stream. Candidate \(i\)'s outcomes are its daily rank-IC
\(X_t\in[-1,1]\) on the days after submission. The null says it has no
edge: \(H_0:\ \mathbb{E}[X_t\mid\mathcal{G}_{t-1}]\le0\). Given
everything known the day before, its expected rank-IC is at most zero.
Each day, the referee stakes a fraction \(\lambda_t\) of the candidate's
capital on a positive outcome,
\[W_t=\prod_{s\le t}\bigl(1+\lambda_s\,X_s\bigr),\qquad \lambda_s\in[0,\lambda_{\max,s}]\ \ \mathcal{G}_{s-1}\text{-measurable},\quad \lambda_{\max,s}<1.\]
The referee fixes \(\lambda_t\) before seeing \(X_t\). The null makes
the expected value of \(X_t\) at most zero. These conditions make \(W\)
a test supermartingale. The cap on \(\lambda\) keeps every product
factor positive. The capital therefore can never be lost outright.

The bet is fair to a null candidate, however cleverly it was chosen,
under one proviso. The proposer's choice must use no information about
the post-submission stream. This condition holds in a live run, whatever
the proposer learned from the past. It may fail in a historical replay.
A model may hold such information if its training window covers the
stream. For the language-model arms, the guarantee therefore applies to
live deployment. The replay results are outcomes
(\S\ref{sec:discussion}).

Every feasible bettor is slow at realistic noise levels because noise
sets the speed. An alternative with conditional mean \(\mu>0\) and
variance \(\sigma^2\) has expected daily log-capital growth of about
\(\lambda\mu-\lambda^2\sigma^2/2\). Growth is largest at the Kelly stake
\(\lambda^*=\mu/\sigma^2\). Capital then reaches \(1/\alpha\) after
roughly \(\ln(1/\alpha)\cdot2\sigma^2/\mu^2\) days. Daily rank-IC on a
500-stock universe has \(\sigma\approx0.12\). For relevant edges
(\(\mu=0.02\)--\(0.03\)), the Kelly stake exceeds the no-bankruptcy cap.
We use the plug-in aGRAPA stake
(\citeproc{ref-waudby2024betting}{Waudby-Smith and Ramdas 2024}), capped
at a fraction \(\phi=0.8\) of that bound:
\[\lambda_t=\mathrm{clip}\Bigl(\frac{\hat m_{t-1}}{\hat v_{t-1}+\hat m_{t-1}^2},\,0,\,\lambda_{\max,t}\Bigr),\]
Here, \(\hat m_{t-1}\) and \(\hat v_{t-1}\) are the running mean and
variance of \(X_s\) over \(s<t\). The cap fraction, not the bettor, sets
the wait (Appendix \ref{app:derivations}).

Autocorrelation changes which null makes the bet fair, so the referee
whitens the stream. A factor with a \emph{marginal} edge of exactly zero
has a positive conditional mean under AR(1) noise after a positive
shock. A bet that conditions on the past exploits those days. The
raw-stream referee admits 6.0\% of such factors over ten years at
\(\rho=0.2\) and 15.4\% at \(\rho=0.4\), against a nominal 5\%. The
referee therefore bets on the \emph{predictably whitened} stream
\[\tilde X_t=X_t-\hat\rho_t X_{t-1},\qquad \hat\rho_t=\text{shrunk, capped estimate from }X_{<t},\qquad \lambda_{\max,t}=\frac{\phi}{1+\hat\rho_t},\]
against \(\mathbb{E}[\tilde X_t\mid\mathcal{G}_{t-1}]\le0\). Every
quantity is \(\mathcal{G}_{t-1}\)-measurable. The supermartingale
argument therefore holds for this null. The theorem gives no
finite-sample bound for the marginal null. Thus, control of the marginal
edge is measured, not proved. Whitening restores false-admission rates
of 1.8, 2.0 and 2.4\% at \(\rho=0,0.2,0.4\) (Table \ref{tbl:ablation},
Appendix \ref{app:derivations}).

Admission certifies an edge. Whether that edge pays for its own trading
is a separate question. The registered threshold \(\delta\) addresses
that question. A factor with edge \(\mu\) earns about \(\kappa\mu\) per
day gross. Here, \(\kappa\) converts one unit of daily IC into return.
It pays \(2c\cdot\mathrm{TO}\) per day for per-side cost \(c\) and
one-way turnover \(\mathrm{TO}\). The break-even edge is
\[\delta=\frac{2c\cdot\mathrm{TO}}{\kappa}.\] \(\kappa\) is the
\emph{average} conversion from daily IC into gross return. We use
\(0.018\) (\S\ref{sec:kappa}). The break-even differs across families by
an order of magnitude because turnover does. At \(c=15\) bp, it is 0.010
for value, 0.012 for liquidity, 0.017 for volatility, 0.031 for momentum
and 0.074 for short-horizon reversal.

One \(\delta\) across these families is therefore a \emph{signal-quality
floor}, not a per-family economic threshold. It was registered as a
single value, \(\delta=0.015\), before the runs. It is the null of the
retirement detector. It defines the realised label of
\S\ref{sec:problem}. It is the execution objective at the shortest hold
(\S\ref{sec:execution}). The execution layer decides whether a certified
factor covers its own costs.

\subsection{Many candidates: online e-BH over a frozen
universe}\label{sec:admission}

The second procedure decides which candidates to admit when many are
available. The rule is Benjamini--Hochberg with capital instead of
\(p\)-values. Its objective is the false-discovery rate over everything
ever admitted. The universe is frozen at \(N_v\) slots per epoch. Each
candidate occupies its own slot with weight \(\gamma_c=w_v/N_v\),
\(\sum_vw_v\le1\). The admitted set on day \(t\) is
\[R_t=\Bigl\{c:\ E_c(t)\ge\frac{1}{k_t^*\,\alpha\,\gamma_c}\Bigr\},\qquad k_t^*=\max\Bigl\{k:\ \#\bigl\{c:E_c(t)\ge\tfrac{1}{k\alpha\gamma_c}\bigr\}\ge k\Bigr\},\]
where \(E_c(t)\) is the candidate's capital, frozen at its admission-day
value (\citeproc{ref-fischer2024online}{Fischer et al. 2024}). More
candidates clearing a bar lowers the bar each faces. Appendix
\ref{app:glossary} works a two-candidate example.

The guarantee requires no assumption about dependence between
candidates. If \(c\) is admitted,
\(\mathbf 1\{c\in R_t\}/|R_t|\le\alpha\gamma_c E_c\), which is
\(\alpha E_c/N_v\) at \(w_v=1\). Taking expectations under each null and
summing over the null candidates gives \(\mathrm{FDR}\le\alpha\)
(\citeproc{ref-wang2022ebh}{Wang and Ramdas 2022}). The guarantee holds
at every stopping time because stopped capital is an e-value. Frozen
values never fall, so admissions are never revoked
(\citeproc{ref-wang2025stopping}{Wang et al. 2025};
\citeproc{ref-deheide2026}{Heide 2026}).

In the runs, the bar is far above the single-candidate \(1/\alpha\). We
set \(\alpha=0.05\) and \(w_v=1\), using one epoch with \(N_v=2000\)
slots. A campaign uses about 490 slots. An expired or retired candidate
keeps its slot, so every admission draws on the same budget. The first
admission requires capital of \(N_v/\alpha\). The \(k\)-th requires
\(N_v/(k\alpha)\).

The bar's height sets the wait. The noise determines that wait.
Combining the bar with the growth rate of \S\ref{sec:betting} gives the
expected wait for the \(k\)-th admission,
\[T\approx\frac{\ln\bigl(N_v/(k\alpha)\bigr)}{\lambda\mu-\lambda^2\sigma^2/2},\qquad T\approx\ln\bigl(N_v/(k\alpha)\bigr)\cdot\frac{2\sigma^2}{\mu^2}\ \ \text{at the Kelly stake.}\]
At the run parameters, the first admission of a factor with edge
\(0.03\) takes about 546 days. At the viability threshold, edge
\(0.015\), it takes 1434 days, versus 1356 for an uncapped Kelly bettor.
The cap costs almost nothing near the threshold (Appendix
\ref{app:glossary} works the three cases). Thus, no better bettor can
shorten the wait near the threshold. A factor whose edge sits at
\(\delta\) needs more than five years of evidence. A campaign lasts
seven to ten.

Two consequences of the rule shape the rest of the paper. A resubmitted
near-copy opens a new candidate with its own slot. Repetition therefore
cannot raise the false-discovery rate above \(\alpha\). It only spends
the proposer's slots. A weak submission does not dilute the other
candidates. The proposer's scarce resources are therefore slots, compute
and calendar time. A bandit allocates these scalar quantities well
(\S\ref{sec:controllers}). The engine can also merge near-duplicates
into a cluster (\citeproc{ref-vovk2021merging}{Vovk and Wang 2021}). The
reported experiments do not merge.

\subsection{Retirement: the e-detector}\label{sec:retirement}

The third procedure decides when a deployed factor's edge has gone. A
desk needs a rule that rarely retires a healthy factor but reacts within
months when the edge is really gone. Real decay is gradual
(\citeproc{ref-mclean2016}{McLean and Pontiff 2016}), so its start has
no ground truth.

The detector plays the betting game of \S\ref{sec:toolbox}
\emph{against} the factor. It uses the viability threshold as the null.
The retirement null says the factor remains viable on the whitened
stream:
\(H_0^{\mathrm{ret}}:\ \mathbb{E}[\tilde X_t\mid\mathcal{G}_{t-1}]\ge\delta_t\)
with \(\delta_t=\delta(1-\hat\rho_t)\). The test bets the other way, on
\(Y_t=\delta_t-\tilde X_t\). Capital grows only while the edge is
demonstrably below the floor. A new e-process \(W^{(j)}\) starts every
day \(j\) after admission because decay can begin on any day. The
statistic is their sum:
\[M_t=\sum_{j\le t}W^{(j)}_t,\qquad \tau_A=\inf\{t:\ M_t\ge A\}.\] The
restarts make the detector responsive. A single e-process started at
admission would dilute new evidence with the factor's healthy period.

The guarantee concerns the average run length. Each \(W^{(j)}\) is a
test supermartingale under the retirement null, so \(M_t-t\) is a
supermartingale. Optional stopping then gives
\(\mathbb{E}_{H_0}[\tau_A]\ge A\) for any \(A\), nonasymptotically and
without i.i.d. assumptions (\citeproc{ref-shin2024edetectors}{Shin et
al. 2024}). The same inequality has the finite-horizon form
\(\mathbb{P}(\tau_A\le t)\le t/A\). It is vacuous over a ten-year
campaign at \(A=1260\). The false-retirement rates below are therefore
simulation results, not consequences of the bound.

The stake targets the decay one wants to catch and is not fitted to the
data. A stake fitted to a healthy factor's drift is near zero. Every
capital then stays near one, \(M_t\approx t\) climbs linearly, and every
healthy factor alarms within a bounded horizon. It is the Kelly bet
against the design alternative ``full decay to zero edge'',
\[\lambda_t=\mathrm{clip}\Bigl(\frac{\delta_t}{\hat\sigma^2_{t-1}+\delta_t^2},\,0,\,\lambda_{\max,t}\Bigr),\]
with \(\hat\sigma^2_{t-1}\) the running variance. Each start then has
negative expected log growth under the alive null. As a result, \(M_t\)
plateaus instead of climbing.

The threshold \(A^*=1260\), five years of trading days, was registered
before the runs. Closed-form economic designs for memory-type charts are
known to fail unchecked (\citeproc{ref-lorenzen1986}{Lorenzen and Vance
1986}; \citeproc{ref-ahmadi2021fallacy}{Ahmadi-Javid and Ebadi 2021}).
The harness therefore checks the delay and false-retirement rate by
simulation. On synthetic decay, the e-detector's median delay is
204--322 days by decay shape. This delay matches a
Monte-Carlo-calibrated CUSUM. The e-detector needs no calibration step.
Its false-retirement rate on ten-year alive streams is 0.5--1.5\% at
\(\rho\le0.2\). At \(\rho=0.4\), the rate is 8.5\%.

Each factor keeps its own detector and threshold when \(K\) factors are
monitored at once. \emph{Error-over-patience}
(\citeproc{ref-dandapanthula2025eop}{Dandapanthula and Ramdas 2025}),
the library-level analogue of the false-alarm rate, is bounded by
\(K/A^*\) under this per-stream rule. Decay shared by a whole family is
read from \emph{family detectors}. Each is the mean of the members'
detector statistics \(M_t\). They inform diagnosis and never retire a
factor by themselves.

\subsection{What an agent in the loop cannot break}\label{sec:validity}

The three guarantees survive an adaptive proposer if the proposer cannot
reach three things. It cannot reach bet sizing, which happens before the
day's outcome is seen. It cannot reach the scoring data, which all
arrives after the candidate was proposed. It cannot change the set of
possible hypotheses within an epoch. Under these conditions, the
proposer's behaviour does not enter the argument.

\textbf{Proposition 1 (validity for any proposal policy).} The
controller may be any policy (scripted, bandit, LLM-driven or
self-modifying) that satisfies four conditions:

\begin{enumerate}
\def\labelenumi{(\roman{enumi})}
\tightlist
\item
  the referee's betting fractions and cluster weights stay
  \(\mathcal{G}_{t-1}\)-measurable;
\item
  each candidate is bet on post-submission data only;
\item
  the library stays frozen within an epoch;
\item
  each candidate \(i\) is chosen with no information about its
  post-submission outcomes beyond \(\mathcal{G}_{s_i-1}\).
\end{enumerate}

Suppose also that each candidate's null is the one its bettor scores.
Every merge must also use weights fixed in advance. The FDR guarantee of
admission then holds at every stopping time. The run-length guarantee of
retirement holds for every deployed factor.

That hypothesis uses the whitened conditional null of
\S\ref{sec:betting}. In words, a null candidate's expected whitened
rank-IC is at most zero, given everything known the day before.

\emph{Proof sketch.} Conditions (i) and (ii) make each candidate's
capital a test supermartingale with respect to \(\mathcal{G}\).
Condition (iv) preserves this property for a candidate selected by the
proposer. The selection carries no information about the increments used
to score the bet. Weights fixed in advance make a merged capital an
e-value. Condition (iii) keeps the hypothesis universe fixed and finite
within each epoch. Thus, stopped e-BH
(\citeproc{ref-wang2025stopping}{Wang et al. 2025}) and dynamic
e-closure (\citeproc{ref-deheide2026}{Heide 2026}) apply. The detector's
guarantee is a per-stream inequality independent of the policy. The
proposition follows from the cited results and is not a new theorem. Its
value is to rule out one possibility: that an endogenous proposal
process breaks validity.

The proposition fails under one named condition, \emph{foretelling}. The
engine guards against it. Foretelling means sizing a bet with knowledge
of where another candidate's stream is about to move. Each bet can then
be fair on its own information. Yet the merged capital can have
expectation above one. The counterexample of
(\citeproc{ref-wang2025stopping}{Wang et al. 2025}) has
\(\mathbb{E}[M]=1.25\). The engine enforces the alignment invariant of
\S\ref{sec:problem}. A unit test reproduces the counterexample and
confirms that the invariant removes it. The invariant cannot verify the
conditional-mean assumption itself. The synthetic world tests that
assumption where it holds by construction.

The proposition also defines the boundary between the agent and the
referee. The trust kernel comprises the referee, the library, the
point-in-time data layer, the append-only decision log and the cost
meter. These frozen components cannot be mounted, patched or intercepted
by the controller. Everything else (probes, memory, allocation and
diagnosis policies) is a swappable agent surface. It may evolve freely.

\section{The execution layer}\label{sec:execution}

The referee certifies a statistic. A book earns money from positions.
The execution layer sits between them. Its one parameter sets how long
to hold a certified ranking. Its one decision is whether to hold the
ranking at all.

\subsection{\texorpdfstring{Persistence, and the edge of an \(h\)-day
hold}{Persistence, and the edge of an h-day hold}}\label{sec:persistence}

How long to hold a ranking depends on how fast its predictive power
fades. The referee does not measure this decay. A ranking \(S_{t-1}\)
known at the close of day \(t-1\) has a \emph{decay curve}
\[IC(k)=\mathbb{E}\bigl[\operatorname{rc}\bigl(S_{t-1},\,r_{t+k-1}\bigr)\bigr],\qquad k=1,2,\dots\]
where \(\operatorname{rc}\) is the cross-sectional rank correlation. The
referee scores \(IC(1)\). \(IC(k)\) for \(k>1\) is the same ranking's
value on the \(k\)-th day of the hold. A flat curve has the same daily
IC at every \(k\). Thus, a daily certificate does not penalise slowness.
A steep curve may clear \(\delta\) yet fail to cover the trading imposed
by its decay.

The objective is the net return of a hold. A \emph{sleeve} is a single
factor's own long--short portfolio. It is long the top fifth and short
the bottom fifth. A sleeve re-formed every \(h\) days earns the curve's
average over the hold. It also pays for re-formation:
\[\mathrm{net}(h)\;=\;\kappa\,\overline{IC}_{1..h}\;-\;2c\,\tau(h),\qquad \overline{IC}_{1..h}=\frac1h\sum_{k=1}^{h}IC(k),\]
where \(\tau(h)\) is the sleeve's one-way turnover per day at that
frequency. Setting \(\mathrm{net}(1)=0\) recovers
\(\delta=2c\,\mathrm{TO}/\kappa\). Thus, \emph{the viability threshold
is the execution objective at the shortest hold}.

\subsection{\texorpdfstring{Estimating \(\kappa\): why the conversion is
a market
constant}{Estimating \textbackslash kappa: why the conversion is a market constant}}\label{sec:kappa}

The objective needs the conversion rate \(\kappa\). We fix it at one
market-level value, \(\bar\kappa=0.018\). The relevant rate is the
\emph{average}, \(\mathbb{E}[\text{gross}]/\mathbb{E}[IC]\). It is
\(0.016\)--\(0.019\) for the four families that carry the book. The
\emph{slope} of daily gross return on daily IC is about three times
larger. It measures the rate on high-dispersion days. The average cannot
be estimated per sleeve. A per-window ratio divides by a 250-day mean IC
whose standard error is the size of a real edge. A first pass that
estimated it per sleeve inverted the shelving decisions (Appendix
\ref{app:derivations}).

\subsection{The decision, and its rate limit}\label{sec:decision}

The decision rule is a deliberately slow grid search. The sleeve's
holding frequency is chosen from five desk frequencies. Each is scored
on the trailing \(W\) days using only inputs known at the decision date.
The search runs on first deployment and then at most once every
\(B=250\) trading days. An accepted change moves at most one grid step.
The sweep over \(B\) finds no consistent winner. The pre-stated rule
\emph{never worst, least churn} therefore selects 250 (Appendix
\ref{app:derivations}).

A factor that cannot pay at any frequency is not traded. If
\(\max_h\widehat{\mathrm{net}}(h)\le0\), the factor stays certified but
is \emph{shelved}. This second threshold is deliberately kept apart from
\(\delta\). The certificate needs one statistic on one daily stream.
Tradability is family-specific because \(\tau(h)\) differs across
families by an order of magnitude.

\subsection{Execution neutrality}\label{sec:neutrality}

\textbf{Proposition 2 (execution neutrality).} Let the execution layer
be any policy mapping the admitted set and data up to \(t-1\) to
positions at \(t\). Then, (i) no guarantee of \S\ref{sec:referee}
depends on it. (ii) Any book can be recomputed from the recorded
decision log without re-running the agent.

\emph{Proof.}

\begin{enumerate}
\def\labelenumi{(\roman{enumi})}
\tightlist
\item
  The e-processes depend on candidate streams and predictable stakes.
  The admission and retirement rules depend on the e-processes. The
  execution layer is downstream of all of them. It enters no bet, weight
  or threshold.
\item
  By construction, the controller's state \(Z_t\) excludes the book
  (\S\ref{sec:controllers}). Therefore, the agent's action sequence
  depends only on the referee-side history. The book is a deterministic
  function of that history.
\end{enumerate}

Part (ii) makes the experiment affordable and auditable. The 540 cells
of \S\ref{sec:results} were recorded once, then re-booked under a new
execution layer at no cost. All 220 LLM cells reproduced their
referee-level metrics exactly.

\section{The controller ladder and the perception
probe}\label{sec:controllers}

The three controllers differ in their capabilities, not their inputs.
All use the same library, budget, and numeric family statistics. A
controller is a policy \(\pi(a_t\mid Z_t)\) over actions
\(\{\text{mine}(f,\theta),\ \text{respond}(\text{alarm}),\ \text{probe},\ \text{idle}\}\).
Its state \(Z_t\) includes family statistics, the referee's diagnostics,
open candidates, and paused families. For the LLM, it also includes
unstructured context and memory. Each rung adds a capability to the rung
below (Table \ref{tbl:ladder}, Appendix \ref{app:amendments}).

\begin{itemize}
\tightlist
\item
  \textbf{C0 round-robin} ignores \(Z_t\). It cycles through families in
  a fixed order and replaces each retired family with a new draw.
\item
  \textbf{C1 discounted UCB} is a non-stationary multi-armed bandit over
  families. Each family's reward is its discounted admission count minus
  retirements per submission.
\item
  \textbf{C2 LLM controller} plays three roles under a fixed scaffold.
  \emph{Direction}: allocate the next period's submissions across
  families using \(Z_t\) and each family's economic profile.
  \emph{Diagnosis} after a retirement: label the cause or escalate.
  \emph{Probe authoring}: write, sandbox-test, and mount a perception
  probe. The probe never enters admission or retirement inference. It
  therefore carries no multiplicity cost.
\end{itemize}

The bandit is a strong allocation baseline. This indicates where to look
for the LLM's contribution. Admissions do not dilute one another
(\S\ref{sec:admission}). The direction role therefore allocates scalar
resources, which a bandit handles well. The LLM has only two inputs
unavailable to the bandit: retirement reasons and family profiles. Any
difference must therefore come from diagnosis and instrument-making.

\section{Experimental design}\label{sec:design}

\subsection{The factorial}\label{sec:factorial}

\begin{figure}[t]
\centering
\scalebox{0.7}{\begin{tikzpicture}[font=\footnotesize,
  q/.style={draw, align=center, text width=50mm, inner sep=5pt, minimum height=15mm},
  arr/.style={-{Stealth[length=2.5mm]}, thick}]
\node[q, fill=orange!15] (tl) at (0,2.6) {\textbf{Intelligence Without Governance}\\ LLM Controller $\times$ Leaky or No Referee\\ (Today's Agentic Miners)};
\node[q, fill=green!12] (tr) at (6.2,2.6) {\textbf{Governed Intelligence}\\ LLM Controller $\times$ Frozen Referee\\ (This Paper's Top Rung)};
\node[q, fill=gray!10] (bl) at (0,0) {\textbf{Neither}\\ Round-Robin $\times$ No Gate\\ (A Fixed IC Checklist)};
\node[q, fill=blue!10] (br) at (6.2,0) {\textbf{Governance Without Intelligence}\\ Script or Bandit $\times$ Frozen Referee\\ (Scripted Proposers)};
\draw[arr] (-3.3,-1.3) -- node[below, pos=0.9]{\small Governance: Who Judges} (9.5,-1.3);
\draw[arr] (-3.3,-1.3) -- node[above, sloped, pos=0.85]{\small Intelligence: Who Proposes} (-3.3,4.0);
\end{tikzpicture}}
\caption{The factorial as a quadrant. Every cell is measured with common random numbers, so the effect of governance (columns) and of intelligence (rows) can be read off separately. The interaction, what the agent adds only when the referee leaks, is a measured quantity.}
\label{fig:quad}
\end{figure}
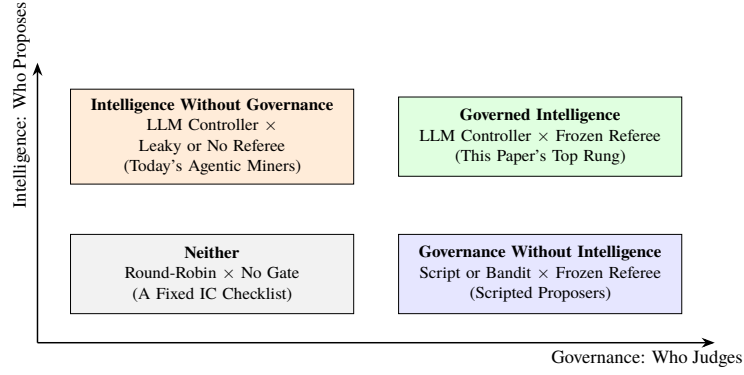

The design crosses who judges with who proposes, and every cell is run.
Four referee arms span governance. Three controllers span intelligence
(Figure \ref{fig:quad}). Three words recur. An \emph{arm} is one factor
setting (a referee, a controller). A \emph{cell} is one run of one arm
combination with one model, start year and seed. A \emph{campaign} is a
cell's walk-forward. It lasts seven to ten years, depending on the start
year.

The governance axis compares one valid referee with three controlled
violations. The \emph{frozen referee} follows the procedure in
\S\ref{sec:referee}. Each \emph{leaky} arm keeps the rest of the
pipeline intact. It breaks one assumption of \S\ref{sec:validity} in the
way a common practice does:

\begin{itemize}
\tightlist
\item
  \textbf{Peeking:} a one-sided \(t\)-test uses the post-submission
  mean. It is checked every day after 60 observations and admits at the
  first \(p\le\alpha\) (the practitioner's rolling \(t>1.65\) rule). It
  breaks optional stopping.
\item
  \textbf{Adaptive threshold:} the e-process remains, but the wealth
  threshold relaxes with candidate age,
  \(\alpha^{-1}e^{-\mathrm{age}/\tau}\). It breaks Ville's fixed
  threshold.
\item
  \textbf{No gate:} every candidate is admitted after a fixed incubation
  of 126 trading days, as today's miners do. It tests nothing.
\end{itemize}

The leaky arms support two contrasts. Comparing a leaky arm with the
frozen arm under one controller measures what the leak lets through.
Comparing two controllers within one leaky arm measures how much
apparent adaptive-controller skill comes from the leak.

The intelligence axis has the three controllers of
\S\ref{sec:controllers}. On real data, the LLM controller uses three
model families. That quadrant is a replay. A language model trained on
the campaign years falls outside condition (iv) of Proposition 1. It
therefore reports what the package achieved and is not an instance of
the guarantee. The referee's design choices were settled through
ablation on synthetic streams before any paid run (Table
\ref{tbl:ablation}, Appendix \ref{app:derivations}).

\subsection{One analysis, three environments}\label{sec:analysis}

Every comparison is a paired difference between two arms that saw the
same world. Within one model family and one start year, arms on the same
seed share common random numbers: the same world and candidate draws.
Their metric difference is one observation, and there are five seeds.
The test is a two-sided paired \(t\)-test on those five differences.
Holm correction covers the tests within one model family. Families are
never pooled. An effect is reported as ``\(k\) of \(n\) families''.
Appendix \ref{app:glossary} states the conventions behind every table.

\emph{The synthetic world isolates the mechanism, because truth is
planted.} It has four families with true-factor proportions
\(0.50/0.25/0.10/0.05\) and AR(1) noise with \(\rho=0.2\). Every
admitted factor lowers the share of true factors among its family's
later candidates. One family has a mid-campaign regime break. A campaign
lasts 1260 days, with four submissions every 21 days. The design is two
LLM families \(\times\) four referee arms \(\times\) three controllers
\(\times\) five seeds.

\emph{The probe environment measures the one agent capability we claim:
writing a diagnostic instrument for a fault nobody listed in advance.}
An episode shows the controller a deployed factor whose series has one
hidden fault. The controller must choose an intervention. The score is
\emph{intervention regret}: the value of the best intervention less the
value of the chosen one. The five fault types are held out, with twenty
episodes each. They are blind spots of the three instruments in the
fixed diagnosis menu. The probe-authoring arm may write its own
instrument, test it in a sandbox, and read it before choosing. Seven
model families are sampled from a written frame (Appendix
\ref{app:amendments}).

\emph{The real-data world tests which results survive ten years of
prices.} It is built point in time from CSI 500 constituents, 2015-01 to
2026-08 (Appendix \ref{app:data}). The four-family library contains the
technical families: momentum, short-horizon reversal, volatility and
liquidity. The nine-family library adds value, size, quality, growth and
investment. Campaigns start in 2016--2019, with five seeds each and four
submissions every 21 days. A candidate's realised label indicates
whether its post-submission mean rank-IC is at least \(\delta=0.015\).

Coverage is complete for the scripted controllers and partial for the
LLM (Table \ref{tbl:coverage}, Appendix \ref{app:amendments}). Three LLM
families run the frozen referee on both libraries. One family runs the
three leaky referees on the four-family library. The other two run its
no-gate arm. Round-robin and bandit baselines run every arm at no cost.
That makes 540 cells: 320 baseline and 220 LLM. Every LLM call is
recorded, so every cell replays at zero cost (Proposition 2).

\subsection{What is fixed before the data, and what is
not}\label{sec:oos}

Every input is sorted by when it was fixed.

\begin{itemize}
\tightlist
\item
  \textbf{Fixed in advance and never tuned:} the level \(\alpha\), the
  viability threshold \(\delta\), the run-length target \(A\), the cap
  fraction, the incubation of the no-gate arm, the parameter grids, and
  the design and analysis plan.
\item
  \textbf{Walk-forward by construction:} every referee verdict, every
  controller action, and every sleeve decision by the execution layer.
\item
  \textbf{In sample:} the conversion constant \(\bar\kappa\) was
  measured once over the whole panel. The execution layer was
  constructed after a first pass over the recorded cells (Appendix
  \ref{app:derivations}).
\end{itemize}

There is no held-out period beyond the walk-forward, and there is one
market. A second market needs data, not model calls. It is the natural
next study.

\subsection{The canonical book}\label{sec:book}

The canonical book is the portfolio construction used for every reported
cell. \textbf{Its numbers are in sample with respect to the execution
layer} and make no out-of-sample claim. The referee-side results are
computed before any book exists (\S\ref{sec:neutrality}). The
construction has four steps.

\begin{itemize}
\tightlist
\item
  \textbf{Sleeves.} Each deployed factor forms a quintile long--short
  sleeve with one unit notional on its own calendar.
\item
  \textbf{Calendar.} The execution layer selects each sleeve's holding
  frequency from five desk frequencies. It scores them on the trailing
  250 days with \(c=15\) bp and \(\bar\kappa=0.018\). A sleeve that
  cannot pay at any frequency is shelved.
\item
  \textbf{Aggregation.} Traded sleeves are combined with equal notional.
  The book's one-way turnover is capped at 25\% per day. An index leg
  hedges its index beta.
\item
  \textbf{Costs.} Costs are 15 bp per side on stocks and 5 bp on the
  hedge. They are charged on position changes.
\end{itemize}

Each book has three reported statistics. The CSI 500 index beta shows
what the hedge leaves. A three-factor alpha follows the spirit of CH-3
(\citeproc{ref-liu2019ch3}{Liu et al. 2019}) and uses Newey--West
\(t\)-statistics (\citeproc{ref-newey1987}{Newey and West 1987}). The
deflated Sharpe ratio (\citeproc{ref-bailey2014dsr}{Bailey and López de
Prado 2014}) discounts for the 108 book-level groups in this study.
Sharpe ratios use 244 trading days a year. Appendix \ref{app:glossary}
defines each statistic operationally.

\section{Results}\label{sec:results}

Three results carry the paper. The referee, not the controller, sets the
number of false admissions (\S\ref{sec:synthetic},
\S\ref{sec:real-referee}, \S\ref{sec:comparator}). The language model
finds more true factors than a script and about as many as a bandit. Its
clearest contribution is writing its own probes (\S\ref{sec:probes},
\S\ref{sec:real-controller}). The certified portfolio trails the ungated
one because a daily certificate arrives late. It also trails because the
family that pays least clears the certificate fastest
(\S\ref{sec:portfolio}). Two controls (\S\ref{sec:controls}) rule out
the two artefacts a reviewer would first suspect.

\subsection{Synthetic factorial: the referee decides false admissions;
the controller does not}\label{sec:synthetic}

\emph{Question.} With truth planted, who determines the number of false
factors admitted? \emph{Answer.} The referee alone (Table
\ref{tbl:synthetic}, Appendix \ref{app:extended}).

False admissions follow the referee arm, regardless of the controller.
Under the frozen referee, both LLM families admit \(0.000\) false
factors per submission. Under the three leaky referees, the rate is
\(0.26\)--\(0.85\) for script, bandit and LLM alike (2 of 2 families).
The guarantee bounds the expected realised false-discovery proportion.
This proportion is \(0\) in every seed under the frozen referee and
\(0.75\)--\(0.95\) under the leaky ones.

The guarantee does not reduce controller yield. Under the frozen
referee, the three controllers have the same yield (\(0.095\)
round-robin, \(0.096\) bandit, \(0.096\) LLM, \(\pm0.002\)). Yield is
lower under the leaky referees. Every admitted factor lowers the share
of true factors among its family's later candidates. A referee that
admits everything spoils the families it draws from. The no-gate arm's
yield is \(0.042\) against the frozen referee's \(0.095\).

An attacking controller extracts false admissions only from a leaky
referee. Hidden retries resubmit unadmitted candidates under new
identifiers rather than drawing fresh ones. They extract \(+0.027\)
false admissions per submission from a peeking referee and \(+0.009\)
from an adaptive-threshold one, and none from the frozen one.

\subsection{Probes: the agent as instrument-maker}\label{sec:probes}

\emph{Question.} Is there a lifecycle task that the agent can do but a
bandit cannot? \emph{Answer.} It can write its own instruments. This has
a clear effect in half of the evaluable model families and causes no
harm in the rest (Table \ref{tbl:probe}).

The comparison uses the fixed diagnosis menu because a bandit can only
choose among listed actions. It cannot write a new one. Every arm sees
the same episodes, so each episode provides a paired contrast. Authored
probes lower intervention regret by \(0.388\), \(0.248\) and \(0.233\)
in three families. Each result is significant after Holm correction over
the six evaluable families. In the other three, the differences are
\(-0.067\), \(-0.021\) and \(+0.074\) and cannot be distinguished from
zero. One of the seven sampled families never produced a probe within
its budget and is inevaluable.

\subsection{Controls}\label{sec:controls}

Two artefacts could produce the LLM's results without skill: what it
remembers and randomness in its decoding. Neither does.

The memory control varies what the LLM may remember. Six memory arms
(none, working, episodic, full, shuffled, oracle) run under the frozen
referee. Each is compared with the full-memory arm. No difference is
detected. After Holm correction, 0 of 10 tests are significant. Every
arm is within \(\pm0.0025\) yield of the full arm.

The decoding control tests whether rerunning the same model on the same
history changes its decisions. A same-policy fork replays a recorded
cell prefix, then lets the same model continue live. In 5 of 5 cells,
the fork reproduces every outcome-level metric of the original.

\subsection{Real data: the referee}\label{sec:real-referee}

\emph{Question.} Does the synthetic result survive ten years of real
prices? \emph{Answer.} Yes. The referee still sets the number of
realised false admissions. On real data, the size of this effect depends
on who proposes (Table \ref{tbl:referee}; Figure \ref{fig:false},
Appendix \ref{app:extended}).

\begin{table}[htbp]
\centering
\caption{Real data, four-family library: realised false admissions per
campaign and realised yield per submission by referee arm and
controller. Means cover four start years \(\times\) five seeds; the LLM
column is one model family.}\label{tbl:referee}
\begin{tabular}{@{}
  >{\raggedright\arraybackslash}p{(\linewidth - 8\tabcolsep) * \real{0.1864}}
  >{\raggedleft\arraybackslash}p{(\linewidth - 8\tabcolsep) * \real{0.1864}}
  >{\raggedleft\arraybackslash}p{(\linewidth - 8\tabcolsep) * \real{0.1017}}
  >{\raggedleft\arraybackslash}p{(\linewidth - 8\tabcolsep) * \real{0.1356}}
  >{\centering\arraybackslash}p{(\linewidth - 8\tabcolsep) * \real{0.3898}}@{}}
\toprule\noalign{}
\begin{minipage}[b]{\linewidth}\raggedright
Referee
\end{minipage} & \begin{minipage}[b]{\linewidth}\raggedleft
False Adm.: Round-Robin
\end{minipage} & \begin{minipage}[b]{\linewidth}\raggedleft
Bandit
\end{minipage} & \begin{minipage}[b]{\linewidth}\raggedleft
LLM (Gemini)
\end{minipage} & \begin{minipage}[b]{\linewidth}\centering
Yield: Round-Robin / Bandit / LLM
\end{minipage} \\
\midrule\noalign{}
Frozen & 11.7 & 9.4 & 11.7 & 0.40 / 0.47 / 0.47 \\
Peeking & 86.2 & 38.4 & 37.4 & 0.41 / 0.50 / 0.55 \\
Adaptive Threshold & 105.2 & 51.0 & 43.5 & 0.39 / 0.44 / 0.48 \\
No Gate & 196.0 & 189.8 & 78.0 & 0.37 / 0.37 / 0.59 \\
\bottomrule
\end{tabular}
\end{table}

The proposer matters only when the referee leaks. Moving from
round-robin to the LLM changes the count by a factor of \(2\)--\(2.5\)
on the leaky arms. It does not change the count on the frozen one. The
bandit halves the count under peeking and adaptive threshold. It leaves
the count unchanged under no gate.

The two kinds of referee also admit different kinds of false factor.
This difference matters for the book. A false admission with a negative
realised edge loses money for as long as it is held. One with a positive
edge below \(\delta\) merely fails to pay for its trading. The frozen
referee's realised false admissions are all of the second kind. None of
its 100 four-family cells has an admitted factor with a negative
realised edge. The detector had retired every false admission before the
campaign ended. Under round-robin, the leaky referees admit 10.8, 32.0
and 90.8 negative-edge factors per campaign. Under the LLM, they admit
1.2, 6.0 and 15.5 (Table \ref{tbl:anatomy}).

The ordering survives every label we tried. The size of the ratio
depends on the label. Moving the threshold from \(0.010\) to \(0.020\)
changes every count. It does not change the ordering (Table
\ref{tbl:labelsens}). Judging each admitted factor only on
post-admission days removes the overlap of \S\ref{sec:problem}. On that
label, the frozen referee's false admissions under round-robin rise from
11.7 to 17.6 per campaign. The leaky referees' false admissions barely
move. Its advantage is then 5--11\(\times\) against 7--17\(\times\) on
the registered label. Under the LLM, it is 2.6--4.9\(\times\) against
3--7\(\times\) (Table \ref{tbl:postlabel}).

Against each family's own break-even, the ratio shrinks to two- to
threefold. Relabelled that way, the frozen referee has 96.0 false
admissions per campaign under round-robin. The three leaky referees have
180.4, 197.4 and 276.2. These counts arise because short-horizon
reversal has a break-even of 0.074 (\S\ref{sec:betting}). The headline
ratio is therefore a statement about the registered label. The frozen
referee certifies a statistical edge. Whether that edge pays is the
execution layer's question (\S\ref{sec:portfolio}).

\subsection{Real data: a legitimate comparator}\label{sec:comparator}

\emph{Question.} The three leaky referees are known shortcuts. What does
the anytime-valid referee add over a patient desk that waits and
corrects for multiplicity? \emph{Answer.} For clearly good or bad
families, it adds a little accuracy and some yield. For many families
with a small positive edge, it gives a fourfold gain in accuracy. That
gain comes from the height of its bar (Table \ref{tbl:comparator},
Appendix \ref{app:extended}).

The comparator is the procedure a careful desk would run without
anytime-valid tools. It tests each candidate once, 500 trading days
after submission (the frozen referee's median wait). It uses a one-sided
\(t\)-test on the whitened stream. It applies Benjamini--Hochberg at
0.05 every time a window closes. It runs as a counterfactual program on
the frozen cell's recorded submission stream.

On the four-family library, the patient desk comes close, at some cost
in yield. Under round-robin, the 500-day comparator admits 16.4 false
factors per campaign against the frozen referee's 11.7. Both are far
from the 86--196 of the leaky referees. Its yield is lower, at 0.358
against 0.403. Its wait is a fixed 500 days against a median of 439. The
frozen referee lets strong factors through early.

On the nine-family library, the comparator is four times less accurate
because its bar is lower. It admits 41.4 false factors against 9.8.
After the decision, 0.33 of what it admits is false, against 0.12. Both
procedures test a zero edge, and both are valid for it. A \(t\)-test on
500 days has power against an edge of \(0.01\). This edge is positive
and below the viability threshold. The fundamental families supply many
such factors. The frozen referee's bar of \(N_v/(k\alpha)\) is far
higher. Therefore, only strong factors clear it within a campaign. Its
selectivity and its wait are the same knob (\S\ref{sec:admission}).

The evidence therefore supports the claim that the judge must be a
valid, multiplicity-corrected statistical procedure. Either valid
procedure admits several times fewer sub-threshold factors than the
shortcuts. Anytime validity then adds three things. The verdict can be
read on any day without choosing a horizon in advance. Strong factors
pass early. The guarantee also holds for a proposer that adapts to the
verdicts. The comparison has one limit. The candidate sequence is the
one the frozen referee produced. Therefore, the comparison isolates the
referee and is not a re-run campaign.

\subsection{Real data: the controller}\label{sec:real-controller}

\emph{Question.} What does a language model add over a script or a
bandit, given a valid referee? \emph{Answer.} It finds more true factors
than a script and about as many as a bandit (Table
\ref{tbl:controller}). It admits fewer false factors than either only
when the referee is broken (Table \ref{tbl:anatomy}).

Here is how to read Table \ref{tbl:controller}. Each entry contains four
signs, one for each start year. Each sign shows the paired difference
between the LLM controller and a baseline. The number in parentheses
counts the starts where that difference is significant after Holm
correction. Tables \ref{tbl:paired} and \ref{tbl:paired9}, Appendix
\ref{app:extended}, report paired effects with standard errors.

\begin{table}[htbp]
\centering
\caption{Real data, frozen referee: signs of the paired difference
between the LLM controller and baseline across the four start years, by
model family. Parentheses show the number of starts significant after
Holm correction within the family. Families are never
pooled.}\label{tbl:controller}
\begin{tabular}{@{}
  >{\raggedright\arraybackslash}p{(\linewidth - 10\tabcolsep) * \real{0.1667}}
  >{\raggedright\arraybackslash}p{(\linewidth - 10\tabcolsep) * \real{0.1667}}
  >{\raggedright\arraybackslash}p{(\linewidth - 10\tabcolsep) * \real{0.1667}}
  >{\raggedright\arraybackslash}p{(\linewidth - 10\tabcolsep) * \real{0.1667}}
  >{\raggedright\arraybackslash}p{(\linewidth - 10\tabcolsep) * \real{0.1667}}
  >{\raggedright\arraybackslash}p{(\linewidth - 10\tabcolsep) * \real{0.1667}}@{}}
\toprule\noalign{}
\begin{minipage}[b]{\linewidth}\raggedright
Library
\end{minipage} & \begin{minipage}[b]{\linewidth}\raggedright
Contrast
\end{minipage} & \begin{minipage}[b]{\linewidth}\raggedright
Metric
\end{minipage} & \begin{minipage}[b]{\linewidth}\raggedright
DeepSeek
\end{minipage} & \begin{minipage}[b]{\linewidth}\raggedright
Gemini
\end{minipage} & \begin{minipage}[b]{\linewidth}\raggedright
GPT
\end{minipage} \\
\midrule\noalign{}
Four-Family & LLM \(-\) Round-Robin & Yield & \(++++\) (4/4) & \(++++\)
(3/4) & \(++++\) (2/4) \\
Four-Family & LLM \(-\) Bandit & Yield & \(++--\) & \(+---\) &
\(----\) \\
Nine-Family & LLM \(-\) Round-Robin & Yield & \(++++\) (4/4) & \(++++\)
(3/4) & \(++++\) (1/4) \\
Nine-Family & LLM \(-\) Bandit & Yield & \(+++-\) & \(+++-\) & \(----\)
(3/4) \\
Four-Family & LLM \(-\) Round-Robin & False Adm. & \(+++-\) (0/4) &
\(-+++\) (0/4) & \(++++\) (0/4) \\
Nine-Family & LLM \(-\) Round-Robin & False Adm. & \(++++\) & \(++++\) &
\(++++\) \\
\bottomrule
\end{tabular}
\end{table}

The LLM finds more true factors than round-robin and no fewer false
ones. Its realised yield is higher in six of six family \(\times\)
library settings. The difference is positive in every start year. Its
false admissions under the frozen referee are at or slightly above
round-robin's (11.7 to 13.3 against 11.7). This occurs because it sends
more submissions to productive families. The increase is
Holm-significant in no start for any family.

The balance against the bandit is level. It depends on the library. On
the four-family library, the LLM is level or behind in most starts. On
the nine-family library, two families lead in three of four starts. One
loses in every start (three of four significant). That library has five
barren families, which punish an allocator that keeps sampling them.

Under leaky referees, the LLM's judgement replaces part of the missing
referee. It reduces realised false admissions relative to round-robin on
all three leaky arms (Holm 4/4 each; three model families of three on
the no-gate arm). Relative to the bandit, it does so only under no gate
(4/4). There, the bandit's admission reward stops discriminating, while
the LLM still reads profiles and retirements.

A period split suggests that the LLM's advantage over the bandit is a
prior that data overtake. We split every campaign at 2020-01-02. This is
the first trading day of the period when the fundamental families lose
their edge in this market. Against the bandit, the LLM leads before the
split and trails afterward in all three model families (Table
\ref{tbl:period}, Appendix \ref{app:extended}). Early in a campaign, the
bandit has no admissions to learn from. The LLM's family profiles are
therefore informative. Later, the bandit has learned what the profiles
said. This is an interpretation, not a test. It does bear on leakage. A
model that had seen the period would be expected to gain most after
2020. The sign is the other way.

\subsection{Real data: the portfolio}\label{sec:portfolio}

\emph{Question.} What does a portfolio of certified factors earn, and
what does the certificate cost an investor? \emph{Answer.} The certified
book does not lose money at the registered cost. Its alpha cannot be
distinguished from zero. Its Sharpe ratio net of costs is below that of
a book that admits everything. The gap has two causes: a horizon
mismatch and the wait. Neither is a verdict on the certificate.

The certified book earns money net of the registered costs, but not
demonstrably. The four-family library has twenty frozen-referee groups
of five seeds. Their net Sharpe is \(+0.33\) to \(+0.50\). Against the
study's 108 groups, the deflated Sharpe ratio is \(0.49\)--\(0.69\).
This is far from the level near one that survives a search of that size.
Alpha is \(+0.25\) to \(+1.13\) bp per day. Its Newey--West \(t\) never
reaches one in any group. Index beta is \(-0.07\) to \(-0.08\). The
nine-family library gives the same picture.

The ungated book ranks first in every start year. Its net Sharpe is
\(+0.59\) to \(+0.77\), versus the frozen referee's \(+0.33\) to
\(+0.50\). The two leaky referees fall between them. The order survives
when every book is restricted to 2020 onwards. It also survives
re-costing (Appendix \ref{app:derivations}).

The first cause of the gap is a horizon mismatch. The ungated book holds
momentum, which a daily certificate cannot see. The per-candidate ledger
records each factor's own sleeve returns for every day it is held. Thus,
the two books can be compared line by line (Table \ref{tbl:ledger}). The
two books earn about the same per day from volatility and liquidity.
Momentum creates the large difference. The ungated book held 892
momentum sleeves at \(+1.69\) bp per day. The frozen book held six.

\begin{figure}
\centering
\includegraphics[width=0.97\linewidth,height=\textheight,keepaspectratio,alt={Persistence and monetisation by family. Left: the decay curve IC(k) of a ranking known at t-1 against the return on day t+k-1, with the viability threshold \textbackslash delta. Right: what a single-factor quintile sleeve nets per year at 15 bp per side when re-formed every h trading days. Reversal nets at most +0.9 bp per day at any frequency on the grid; momentum is not certifiable at one day and pays only when held a month or longer.}]{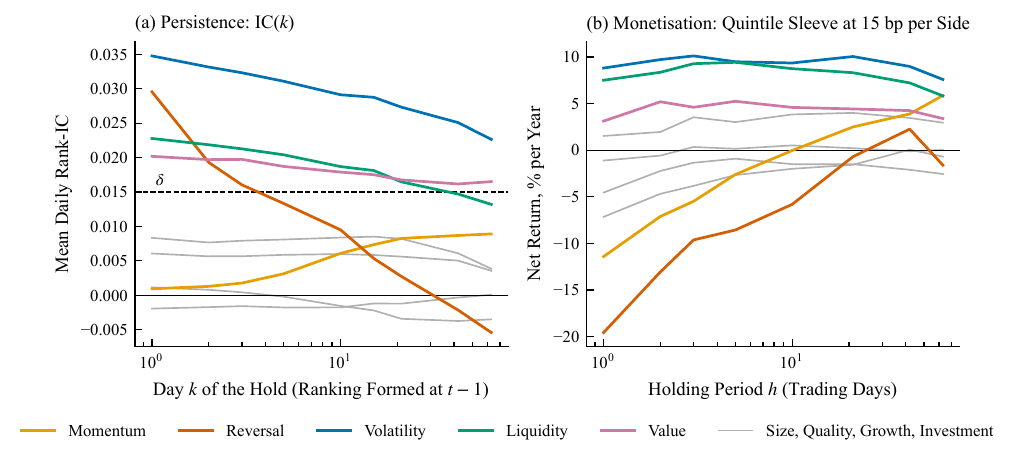}
\caption{Persistence and monetisation by family. Left: the decay curve
\(IC(k)\) of a ranking known at \(t-1\) against the return on day
\(t+k-1\), with the viability threshold \(\delta\). Right: what a
single-factor quintile sleeve nets per year at 15 bp per side when
re-formed every \(h\) trading days. Reversal nets at most \(+0.9\) bp
per day at any frequency on the grid; momentum is not certifiable at one
day and pays only when held a month or longer.}\label{fig:curves}
\end{figure}

The referee is right about momentum on its own terms. It is wrong for
the book. Momentum's daily rank-IC is \(0.001\), an order of magnitude
below \(\delta\). It therefore cannot be certified within a campaign.
But momentum is a slow factor in this market. Its information
coefficient \emph{rises} with the holding day, from \(0.001\) at one day
to \(0.009\) at sixty-three. It pays only when held a month or longer
(Figure \ref{fig:curves}). A library that certifies one-day predictive
power cannot certify momentum. A library that certifies nothing includes
it incidentally.

\begin{table}[htbp]
\centering
\caption{Per-family sleeve economics under the canonical book, using the
round-robin controller and four-family library. Results pool five seeds
from the 2016 start. They report sleeves ever held, the share shelved,
and each family's daily earnings while held, \textbf{gross of their own
trading costs}. Re-formed daily, a reversal sleeve nets \(-19.6\)\% per
year (Figure \ref{fig:curves}).}\label{tbl:ledger}
\begin{tabular}{@{}
  >{\raggedright\arraybackslash}p{(\linewidth - 4\tabcolsep) * \real{0.2963}}
  >{\centering\arraybackslash}p{(\linewidth - 4\tabcolsep) * \real{0.3519}}
  >{\centering\arraybackslash}p{(\linewidth - 4\tabcolsep) * \real{0.3519}}@{}}
\toprule\noalign{}
\begin{minipage}[b]{\linewidth}\raggedright
Family
\end{minipage} & \begin{minipage}[b]{\linewidth}\centering
Frozen: Held / Shelved / bp per Day
\end{minipage} & \begin{minipage}[b]{\linewidth}\centering
No Gate: Held / Shelved / bp per Day
\end{minipage} \\
\midrule\noalign{}
Volatility & 473 / 1\% / \(+4.05\) & 418 / 2\% / \(+4.06\) \\
Liquidity & 314 / 0\% / \(+3.79\) & 555 / 17\% / \(+3.96\) \\
Short-Horizon Reversal & 539 / 30\% / \(+2.26\) & 475 / 41\% /
\(+1.50\) \\
Momentum & \textbf{6} / 83\% / \(-4.35\) & \textbf{892} / 44\% /
\(+1.69\) \\
\bottomrule
\end{tabular}
\end{table}

\begin{figure}
\centering
\includegraphics[width=0.97\linewidth,height=\textheight,keepaspectratio,alt={Cumulative net return of the canonical book, 2016 start, seed-averaged (four-family library; left the round-robin controller, right the LLM controller). The shaded band is the frozen referee's wait. The ungated book begins compounding in its first year while the certified book is still empty, and is below water until year six. The two slopes are similar once both are invested.}]{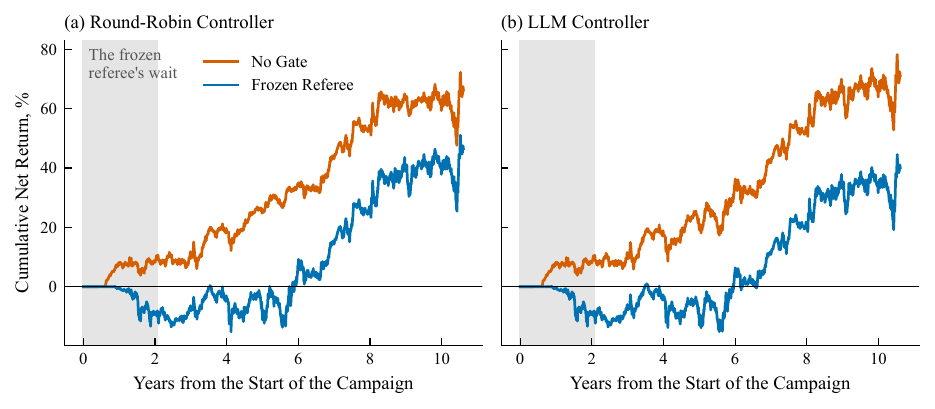}
\caption{Cumulative net return of the canonical book, 2016 start,
seed-averaged (four-family library; left the round-robin controller,
right the LLM controller). The shaded band is the frozen referee's wait.
The ungated book begins compounding in its first year while the
certified book is still empty, and is below water until year six. The
two slopes are similar once both are invested.}\label{fig:cum}
\end{figure}

The second cause is the wait. The wait counts the trading days between a
factor's submission and admission. For an admitted true factor, its
median is about 500 trading days under the frozen referee. It is 122 to
214 under the leaky referees (Figure \ref{fig:delay}, Appendix
\ref{app:extended}). Figure \ref{fig:cum} shows the wait's effect on a
book. The ungated book begins compounding in its first year while the
certified book remains empty. Once both are invested, the two slopes are
similar. The gap opens during the wait and is not earned afterwards.

The wait is expensive because it is paid in regimes, not days. In the
nine-family library, 99--100\% of certified holding windows are centred
after 2020-01-02. The corresponding range is 65--100\% by family for the
ungated pipeline (Table \ref{tbl:regime}, Appendix \ref{app:extended}).
In this market, the edges of the fundamental families value, quality and
growth disappeared after 2020. The certified book earns nothing from
exactly those families. Waiting for volatility and liquidity costs
almost nothing because their edge is stable. A slow certificate is cheap
when the edge persists and expensive when it decays. A descriptive
decomposition of the Sharpe gap agrees. Earliness is its largest term on
the three longer campaigns (Table \ref{tbl:decomp}).

Two counterfactual books test this account (Table \ref{tbl:cfbooks},
Appendix \ref{app:extended}). Both use the round-robin frozen cell's own
candidate sequence and the world's own accounting. The frozen book is
reproduced day by day in 20 of 20 cells.

The first counterfactual switches retirement off. It shows that
retirement is not what the certified book loses. Every admitted factor
is held until the campaign ends. Net Sharpe changes by \(+0.00\) to
\(+0.01\) across the four start years. Some real-data retirements are
false on the realised label. Reversal factors were retired after about
838 days. Their edge had sagged to \(0.0145\) while held. It averaged
\(0.034\) over the rest of the campaign (Table \ref{tbl:retire}). Those
false retirements cost the book almost nothing.

The second counterfactual replaces the frozen referee with the 500-day
comparator. It shows that a desk with the same patience builds a better
book without being earlier. The comparator's net Sharpe is \(+0.59\) to
\(+0.65\), against \(+0.40\) to \(+0.49\) for the frozen book. This is a
paired gain of \(+0.12\) to \(+0.24\) in every start year. Its factors
are not admitted sooner. The mean wait is 500 days against 510. The mix
differs. The comparator admits fewer reversal factors, 72.0 against
87.2. It admits about five momentum factors, while the frozen referee
admits almost none, 4.8 against 0.5. The highest daily rank-IC reaches
the bar fastest, so the bar favours the family with the worst net
economics. A referee with the same patience does not pay that cost.

Certification still buys one thing the Sharpe ratio does not show. The
execution layer never sees the referee. It shelves 13\% of what the
frozen referee admitted, against 26\% of what the ungated pipeline
deployed. More of what the referee admits is worth trading. A portfolio
Sharpe ratio is therefore not a verdict on certification in either
direction. It combines a referee with an execution layer that is blind
to it. The same holds for LLM investment agents evaluated as a whole:
the capability leader is not the Sharpe leader, and the process must be
diagnosed rather than a period's return ranked
(\citeproc{ref-qu2026clqt}{Qu and Chen 2026}).

\section{Discussion and limitations}\label{sec:discussion}

The experiments support the division of labour and show its cost. The
judge must be a statistical procedure the agent cannot touch. Every
leaky referee admitted several times more false factors than the frozen
one under every proposer. No proposer closed that gap. A fixed-horizon
test with a multiplicity correction is also a statistical judge. Anytime
validity adds a verdict that can be read on any day. Its selectivity is
paid for in waiting.

The proposer can be anything. A language model earns its place through
jobs a bandit cannot do. It does not allocate consistently better than a
bandit. It adds three things. It reads why a factor was retired. It
writes an instrument for a fault the menu does not cover. It also holds
the line when the referee leaks.

The split costs time. This time reflects an information bound rather
than an engineering gap. An admitted true factor waits a median of about
five hundred trading days in this market. The wait scales like
\(\ln(N_v/(k\alpha))\cdot2\sigma^2/\mu^2\). At this noise level, no
bettor can certify a factor at the threshold in under five years. Only a
less noisy statistic or a lower bar shortens the wait.

The comparator shows a second price. A desk with the same patience and a
fixed-horizon test builds a better book. Its admissions are less
concentrated in short-horizon reversal. An anytime-valid bar rewards the
statistic that is strongest per day. The strongest daily statistic in
this library is the hardest to trade. The design change is to certify
the horizon that is traded, not the daily one.

The evidence has clear edges, and each edge bounds one claim:

\begin{itemize}
\tightlist
\item
  \emph{The sample.} The sample covers one market and four start years
  that share most of their data. There is no held-out period beyond the
  walk-forward.
\item
  \emph{The label.} The registered label leans towards the frozen
  referee. It is not planted truth and overlaps the admission window.
  Judging each factor only after its admission lowers the headline ratio
  by about a third. The synthetic factorial has no overlap and carries
  the claim about false admissions.
\item
  \emph{The comparisons.} The LLM runs the leaky referees only on the
  four-family library. The fixed-horizon comparator is a counterfactual
  referee on the frozen referee's candidate sequence. It is not an arm
  with its own campaign. The diagnosis role of the LLM controller is not
  ablated on its own. The controller contrasts compare packages, since
  the LLM also reads profiles and memory, and on real data they are
  replay outcomes, not forecasts.
\item
  \emph{The guarantee.} The proof controls the conditional null on the
  whitened stream. The marginal-null guarantee under autocorrelation is
  measured, not proved. The cluster merge is not exercised.
\item
  \emph{Retirement.} Retirement is calibrated on synthetic decay. Its
  guarantee concerns the average run length. This gives no usable bound
  over one campaign.
\item
  \emph{The portfolio.} The portfolio results support no claim of
  investability. Alpha \(t\)-statistics are below one in every certified
  four-family group. The cost model includes no borrow fee. The book is
  an instrument, not a strategy (Appendix \ref{app:derivations}).
\end{itemize}

\section{Conclusion}\label{sec:conclusion}

Judging must be governed by a statistical procedure that the agent
cannot touch. Proposing and instrument-making are where the agent's
intelligence has measurable, bounded value. This division of labour is
governed self-evolution, the framework we have proposed and measured.
The agent's surface evolves freely. A frozen referee decides what enters
the portfolio. Who judges determines how many false factors an always-on
miner certifies. No proposer substitutes for the judge.

The split is testable because three objects remain separate and
auditable: the certificate for the statistic, the measurement of its
persistence, and the execution decision. The certificate is judged on
the referee's own daily stream. Persistence and execution are measured
afterwards, on returns the certificate never saw. A portfolio Sharpe
ratio mixes all three. That is why it is evidence neither for nor
against a certificate. It also explains the next design question: the
certified horizon should match the traded one.

\section*{Data and code availability}\label{data-and-code-availability}
\addcontentsline{toc}{section}{Data and code availability}

The code, run manifests, prompts, and recorded model calls are available
from the corresponding author on request (deepgroundingai@gmail.com).
The authors cannot redistribute the market data because they license
them from a commercial data vendor.

\renewenvironment{CSLReferences}[2]{\begin{list}{}{\setlength{\itemindent}{-1\cslhangindent}\setlength{\leftmargin}{\cslhangindent}\setlength{\parsep}{0pt}\setlength{\itemsep}{0.5\baselineskip}}}{\end{list}}

\section*{References}\label{references}
\addcontentsline{toc}{section}{References}

\protect\phantomsection\label{refs}

\clearpage
\begin{appendices}
\AtBeginEnvironment{tabular}{\small}
\renewcommand{\arraystretch}{0.9}
\setlength{\tabcolsep}{3.5pt}
\renewcommand{\floatpagefraction}{0.85}
\renewcommand{\topfraction}{0.9}
\renewcommand{\bottomfraction}{0.8}
\renewcommand{\textfraction}{0.08}
\setcounter{totalnumber}{4}
\setcounter{topnumber}{3}
\setcounter{bottomnumber}{2}
\setlength{\floatsep}{6pt plus 2pt minus 2pt}
\setlength{\textfloatsep}{8pt plus 2pt minus 3pt}
\setlength{\intextsep}{6pt plus 2pt minus 2pt}
\raggedbottom
\makeatletter
\setlength{\@fptop}{0pt}
\setlength{\@fpsep}{8pt plus 2pt}
\setlength{\@fpbot}{0pt plus 1fil}
\makeatother

\section{Definitions, conventions and worked
calculations}\label{app:glossary}

This appendix supports the body on three things. It defines the terms
the body names before it builds them. It states the conventions behind
every reported number. It works three calculations the body quotes.
Nothing here is a new result.

\emph{Terms at first use.} An \emph{e-process} is a candidate's betting
capital read over time. Under the null it is a test supermartingale, so
its value at any stopping time is a fair quantity, an \emph{e-value}
(\S\ref{sec:toolbox}). \emph{Online e-BH} is the Benjamini--Hochberg
procedure applied to those e-values as they arrive, over a universe of
hypotheses fixed in advance (\S\ref{sec:admission}). An
\emph{e-detector} is a sum of e-processes restarted every day, used as a
changepoint chart (\S\ref{sec:retirement}). A \emph{leaky referee} is a
referee that breaks one assumption of Proposition 1 in the way a common
practice does (\S\ref{sec:factorial}). \emph{Planted truth} means that
the synthetic world knows which of its factors have an edge, so every
admission is labelled true or false without a proxy
(\S\ref{sec:analysis}). A \emph{perception probe} is a small program the
controller writes to measure a suspected fault in a deployed factor. Its
output informs the controller's diagnosis and never enters a statistical
test (\S\ref{sec:controllers}). A \emph{sleeve} is one factor's own
quintile long--short portfolio. A sleeve is \emph{shelved} when it
cannot cover its costs at any holding frequency (\S\ref{sec:execution}).

\emph{Candidate clocks.} A candidate is open, deployed or retired
(\S\ref{sec:problem}). A candidate that leaves the open state without
admission has \emph{expired}. It keeps its slot (\S\ref{sec:admission}).
A \emph{false retirement} is an alarm of the retirement detector raised
while the deployed factor's edge is still at least \(\delta\). The
\emph{run length} is the number of trading days from the detector's
start to its alarm. The guarantee of \S\ref{sec:retirement} bounds the
expected run length to a false retirement from below by \(A^*=1260\)
days. Over one campaign that bound is not informative. The
false-retirement rates reported are therefore simulation results.

\emph{Admission bookkeeping, with a two-candidate example.} An
\emph{epoch} is a period in which the library is frozen. The reported
runs use one epoch. A \emph{slot} is one of the \(N_v=2000\) places in
the registered universe. Every submitted candidate takes one slot and
keeps it, whether it is admitted, expires or is retired. With \(w_v=1\)
every slot has weight \(\gamma_c=1/N_v\). The bar for the \(k\)-th
admission at level \(\alpha=0.05\) is \(N_v/(k\alpha)\). It is 40,000
for the first admission and 20,000 for the second. Suppose candidate A
has capital 45,000 and candidate B has 25,000. A alone clears the first
bar. Both clear the second bar, so \(k^*=2\) and both are admitted. If B
had 15,000 instead, only A would be admitted and B would keep betting.
The rule lowers the bar as more candidates clear it. It never raises the
bar.

\emph{The wait, worked.} At the Kelly stake the expected wait for the
\(k\)-th admission is
\(T\approx\ln\bigl(N_v/(k\alpha)\bigr)\cdot2\sigma^2/\mu^2\)
(\S\ref{sec:admission}). For the first admission
\(\ln(40{,}000)\approx10.6\). Daily rank-IC has \(\sigma\approx0.12\).
At edge \(\mu=0.03\), \(2\sigma^2/\mu^2=32\), so \(T\approx339\) days.
At the viability threshold \(\mu=0.015\), \(2\sigma^2/\mu^2=128\), so
\(T\approx1356\) days. The capped bettor as run, with the stake at the
cap of 0.8, takes 546 and 1434 days in the same two cases. A universe
sized to the campaign's 490 slots would lower every bar by
\(\ln(2000/490)\). That is worth 72 days at edge \(0.03\) and 191 days
at \(0.015\).

\emph{The four threats and their answers.} The body names four threats
(\S\ref{sec:problem}). Each has one answer.

\begin{enumerate}
\def\labelenumi{\arabic{enumi}.}
\tightlist
\item
  \textbf{Optional stopping.} The capital is a test supermartingale. It
  may therefore be read every day and stopped by any rule based on the
  past, and the false-discovery bound holds at the stopping time
  (\S\ref{sec:toolbox}).
\item
  \textbf{Near-copy resubmission.} Every copy takes its own slot under
  online e-BH. The bound stays at \(\alpha\). Repetition spends the
  proposer's slots and nothing else (\S\ref{sec:admission}).
\item
  \textbf{Verifier-in-the-loop adaptation.} The referee sizes each bet
  before the day's outcome and scores only data that arrived after
  submission. What the controller learns about the referee changes which
  candidates it submits, not the referee's error (Proposition 1).
\item
  \textbf{Knowledge-cutoff leakage.} The referee does not address it.
  Proposition 1 assumes that the proposer holds no information about the
  post-submission stream. A live run satisfies this. A historical replay
  by a model trained on the campaign years may not. The language-model
  arms' real-data results are therefore \emph{replay outcomes}: what the
  package achieved on that history, not an instance of the guarantee.
\end{enumerate}

\emph{The analysis unit and the reporting conventions.} The unit of
analysis is a seed. Two arms on the same seed, model family and start
year share common random numbers, and their metric difference is one
observation. A reported ``\(\pm\)'' is the standard error over the five
seeds. The test is a two-sided paired \(t\)-test on the five
differences. Holm correction is applied within one model family. On real
data that family has 24 tests: four starts \(\times\) two contrasts
\(\times\) three metrics. In the synthetic world it has 8: four referees
\(\times\) two metrics. An effect is reported as ``\(k\) of \(n\)
families''. In the sign tables, each entry has one sign per start year.
The number in parentheses counts the starts significant after Holm
correction. No parentheses means none.

\emph{Portfolio statistics, operationally.} A \emph{group} is the five
seeds of one referee, controller, library and start year. The study has
108 groups. The net Sharpe ratio is the mean daily net return over its
standard deviation, annualised with 244 trading days. The deflated
Sharpe ratio (\citeproc{ref-bailey2014dsr}{Bailey and López de Prado
2014}) discounts that ratio for the 108 groups compared. Alpha is the
intercept of a regression of daily net return on market, size and
book-to-price legs built from the panel, in the spirit of CH-3
(\citeproc{ref-liu2019ch3}{Liu et al. 2019}), with Newey--West
\(t\)-statistics at lag 5 (\citeproc{ref-newey1987}{Newey and West
1987}). It is quoted in basis points per day. Index beta is the slope on
the CSI 500 index return.

\emph{The factorial in one schema.} Each environment has a unit, an
outcome and a contrast.

\begin{itemize}
\tightlist
\item
  \textbf{Synthetic world.} Unit: a cell (one arm combination, one model
  family, one seed). Outcomes: realised yield per submission and
  realised false admissions per campaign, labelled against planted
  truth. Contrasts: four referee arms \(\times\) three controllers, two
  LLM families, five seeds.
\item
  \textbf{Probe environment.} Unit: an episode (one deployed factor with
  one hidden fault). Outcome: intervention regret. Contrast: the
  authored probe against the fixed diagnosis menu on the same 100
  episodes, per model family, seven families.
\item
  \textbf{Real data.} Unit: a cell. Outcomes: the realised label at
  \(\delta=0.015\) and the canonical book's statistics. Contrasts: the
  same arms, four start years \(\times\) five seeds, with LLM coverage
  as in Table \ref{tbl:coverage}.
\end{itemize}

\section{Derivations and ablation details}\label{app:derivations}

\emph{Bettor ablation.} The bettor choice is immaterial at these noise
levels. The cap fraction sets the wait. At \(\mu=0.03\) and
\(\sigma=0.12\), the Kelly stake exceeds any no-bankruptcy cap. Thus,
aGRAPA, the Gaussian-efficient bettor
(\citeproc{ref-martinez2026gaussian}{Martinez-Taboada and Ramdas 2026})
and a capped oracle all sit at the cap. A cap fraction of 0.8 accepts a
\(\mu=0.03\) factor in a median of 162 days (\(\rho=0\)) or 190 days
(\(\rho=0.2\)). At a fraction of 0.5, the median is 238/284, with null
false acceptance of 1.9/3.4\%.

\emph{Holding grid.} The grid covers the five frequencies a desk trades:
weekly, fortnightly, monthly, bi-monthly and quarterly. It is scored
using the average of the decay curve. It does not use the practitioner's
``IC at horizon \(h\)''. That measure rises mechanically with \(h\)
because averaging returns lowers noise. The design excludes daily
formation. This costs under 1\% per year for the two families flat
enough to favour it.

\emph{Conversion estimator.} A conversion rate estimated per sleeve is
too noisy to guide shelving. The first canonical pass estimated it per
sleeve. This reversed the cost curve. It shelved 77\% of volatility and
48\% of liquidity sleeves, against 32\% of reversal. The denominator
caused this reversal. It is a 250-day mean IC whose standard error
(\(0.006\)--\(0.013\)) matches the size of a real edge. With the market
constant, the cells shelve 0\% of liquidity, 2\% of volatility, 30\% of
reversal and 89\% of momentum.

\emph{Cool-down sweep.} The pre-stated rule \emph{never worst, least
churn} selected the cool-down \(B=250\), not Sharpe. Dynamic variants
with \(B\in\{5,10,21,63,250\}\) differ by 0.4--3.8\% per year by family.
No variant wins consistently. \(B=250\) exceeds fixed monthly in four of
five families. It is the only variant that turns reversal positive.

\emph{The conversion constant is in sample.} \(\bar\kappa=0.018\) is the
one input measured over the whole sample. Across the half-sample band
\(0.013\)--\(0.026\), the traded-or-shelved decision flips for
3.3--7.2\% of the 51,531 calibrated sleeves. Almost all flips are in
reversal (14--19\% of its sleeves). The books were not re-booked across
the band.

\emph{Cost sensitivity.} Table \ref{tbl:cost} re-costs the 2016
round-robin books with fixed positions and scaled stock-side cost. The
certified book's one-way turnover is 3.4\% per day, against the ungated
book's 2.2\%. Thus, ``the certified book does not lose money'' applies
at the registered cost level.

\begin{table}[htbp]
\centering
\caption{Cost sensitivity of the two books (five-seed means). The
certified book crosses zero between 30 and 50 bp a side. The ungated one
does not.}\label{tbl:cost}
\begin{tabular}{@{}lll@{}}
\toprule\noalign{}
Cost per Side & Certified Net SR & Ungated Net SR \\
\midrule\noalign{}
15 bp (registered) & \(+0.40\) & \(+0.73\) \\
25 bp & \(+0.25\) & \(+0.61\) \\
30 bp & \(+0.17\) & \(+0.54\) \\
50 bp & \(\mathbf{-0.12}\) & \(+0.29\) \\
\bottomrule
\end{tabular}
\end{table}

\emph{Accounting conventions.} Reviewers questioned two conventions of
the registered book. Neither changes the comparison (Table
\ref{tbl:accounting}). \emph{Carry} holds a halted position at its last
price and credits the whole gap on the reopening day. It does not book a
missing return as zero. \emph{Lag} executes the target book decided on
day \(t\) on day \(t+1\). It does not execute at the signal's own close.
Both are re-booked at zero cost from the recorded holding intervals. The
canonical re-booking reproduces every recorded book to \(10^{-9}\).

\begin{table}[htbp]
\centering
\caption{The 2016 round-robin books under four accounting conventions
(five-seed means with standard errors, 244-day annualisation). The
ungated book's lead is \(0.34\)--\(0.35\) under every
convention.}\label{tbl:accounting}
\begin{tabular}{@{}llrr@{}}
\toprule\noalign{}
Book & Convention & Net Sharpe & Cumulative Net Return \\
\midrule\noalign{}
Certified & Registered & \(+0.40 \pm 0.03\) & \(+46.6\)\% \\
Certified & Carry & \(+0.40 \pm 0.03\) & \(+47.1\)\% \\
Certified & One-Day Lag & \(+0.37 \pm 0.03\) & \(+43.5\)\% \\
Certified & Carry and Lag & \(+0.38 \pm 0.03\) & \(+44.1\)\% \\
Ungated & Registered & \(+0.73 \pm 0.02\) & \(+66.3\)\% \\
Ungated & Carry & \(+0.75 \pm 0.02\) & \(+67.7\)\% \\
Ungated & One-Day Lag & \(+0.71 \pm 0.01\) & \(+64.1\)\% \\
Ungated & Carry and Lag & \(+0.72 \pm 0.01\) & \(+65.5\)\% \\
\bottomrule
\end{tabular}
\end{table}

\emph{Instrument properties.} The book is a measuring instrument. The
paper gives no certificate of tradability. Short selling is assumed for
a quintile of individual A-shares, without a borrow list or borrow fee.
The hedge leg uses index futures. A missing return is booked as zero
(12.5\% of member-stock days in 2015, about 0.2\% from 2019). Neither
capacity nor drawdown is modelled. Measured maximum drawdown is 21--25\%
for the certified book and 20--21\% for the ungated one.

\emph{Referee ablations.} The adopted referee whitens the stream. It
uses the predictable bankruptcy bound and estimates \(\rho\). Ablations
on synthetic streams settled each choice before any paid run (Table
\ref{tbl:ablation}). Whitening restores nominal control at every
autocorrelation tested. The predictable bound recovers most of the power
lost under a worst-case bound. Estimating \(\rho\) beats knowing it at
\(\rho=0.4\). This advantage arises because the shrunk estimate
under-whitens, trading a within-bound validity error for power.

\begin{table}[htbp]
\centering
\caption{Referee ablations use synthetic AR(1) streams at \(\rho=0\) /
\(0.2\) / \(0.4\). This table uses cap fraction 0.5. The referee runs at
0.8. The experiment uses ten-year null streams at \(\mu=0\) and
alternative streams at \(\mu=0.03\). Alive streams cover 2520 days with
run-length target 1260. The nominal level is 5\%. Against a
Monte-Carlo-calibrated CUSUM, the adopted e-detector's median delay is
204--322 days by decay shape, against CUSUM's
181--337.}\label{tbl:ablation}
\begin{tabular}{@{}
  >{\raggedright\arraybackslash}p{(\linewidth - 8\tabcolsep) * \real{0.2800}}
  >{\centering\arraybackslash}p{(\linewidth - 8\tabcolsep) * \real{0.1800}}
  >{\centering\arraybackslash}p{(\linewidth - 8\tabcolsep) * \real{0.1800}}
  >{\centering\arraybackslash}p{(\linewidth - 8\tabcolsep) * \real{0.1800}}
  >{\centering\arraybackslash}p{(\linewidth - 8\tabcolsep) * \real{0.1800}}@{}}
\toprule\noalign{}
\begin{minipage}[b]{\linewidth}\raggedright
Stream Transformation
\end{minipage} & \begin{minipage}[b]{\linewidth}\centering
Null False Acceptance, \%
\end{minipage} & \begin{minipage}[b]{\linewidth}\centering
Median Days to Accept
\end{minipage} & \begin{minipage}[b]{\linewidth}\centering
Detector False Alarms on Alive Streams, \%
\end{minipage} & \begin{minipage}[b]{\linewidth}\centering
Step-Decay Delay, Median Days
\end{minipage} \\
\midrule\noalign{}
Raw daily rank-IC (conditional null) & 2.1 / 6.0 / 15.4 & 228 / 227 /
227 & 1.5 / 11.5 / 46 & 213 / 181 / 195 \\
Whitened, worst-case bound & 1.1 / 1.3 / 1.8 & 336 / 378 / 462 & 1.0 /
0.0 / 7.0 & 250 / 264 / 316 \\
Whitened, predictable bound (adopted) & 1.8 / 2.0 / 2.4 & 238 / 284 /
384 & 1.5 / 0.5 / 8.5 & 218 / 234 / 308 \\
Whitened with the true \(\rho\) (oracle) & 2.1 / 1.3 / 0.9 & 228 / 344 /
536 & 1.5 / 0.5 / 6.5 & 213 / 246 / 333 \\
\bottomrule
\end{tabular}
\end{table}

\emph{Family removal.} Removing each family in turn from the ungated
book shows that momentum helps through diversification, not return.
Dropping momentum costs \(+0.11\) to \(+0.18\) net Sharpe while raising
cumulative net return in all four starts. A momentum-only book earns
\(+0.22\) to \(+0.29\). Dropping the certified reversal family raises
the book's Sharpe by \(+0.01\) to \(+0.14\).

\emph{Selection.} We chose the execution layer after first examining the
recorded cells. One fact limits how much this choice can explain. Every
construction tried (A10b, A11, A11b, A11c in Table \ref{tbl:amendments})
ranks the ungated book above the certified one. Neither execution
parameter was chosen on Sharpe. The deflated Sharpe ratio treats the 108
final book groups, not the constructions, as trials. The reported
deflation therefore understates the search.

\emph{Newey--West bandwidth.} At Bartlett bandwidths 5, 8, 21 and 63,
the certified book's alpha \(t\)-statistics are \(0.49\), \(0.49\),
\(0.51\) and \(0.54\). Those for the ungated book are \(1.68\),
\(1.71\), \(1.75\) and \(1.86\). The inference therefore does not depend
on the bandwidth. We keep lag 5.

\emph{Benchmark construction.} The regression uses a characteristic
three-factor model on our own panel, not the published CH-3 benchmark
(\citeproc{ref-liu2019ch3}{Liu et al. 2019}). SMB and HML are
long--short quintiles of \(-\log\) circulating capitalisation and
book-to-price over the full universe. The benchmark legs are gross,
while the dependent variable is net. This difference is expected to
lower the estimated alpha.

\emph{Rank-threshold retirement.} Declaring the top \(k\) streams at
\(A^*/k\) satisfies the error-over-patience bound but is unusable. A
healthy factor's statistic plateaus near \(1/(1-\text{decay})\), with
median \(\approx75\) on the whitened stream. Yet \(A^*/k\) is 63--126
for \(k=10\)--\(20\). The rule therefore declares an all-alive library
of 20 en bloc within a year.

\section{Extended results}\label{app:extended}

\begin{figure}[tbp]
\centering
\includegraphics[width=0.8\linewidth]{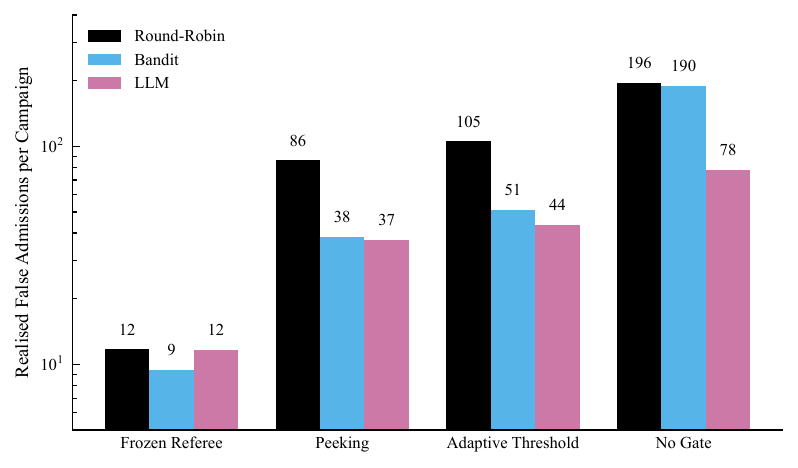}
\caption{Realised false admissions per campaign by referee arm and controller (real data, four-family library; means over four start years $\times$ five seeds; log scale). The referee sets the count; the controller moves it by a factor of 2--2.5, and only when the referee leaks.}
\label{fig:false}
\end{figure}

\begin{figure}[tbp]
\centering
\includegraphics[width=0.8\linewidth]{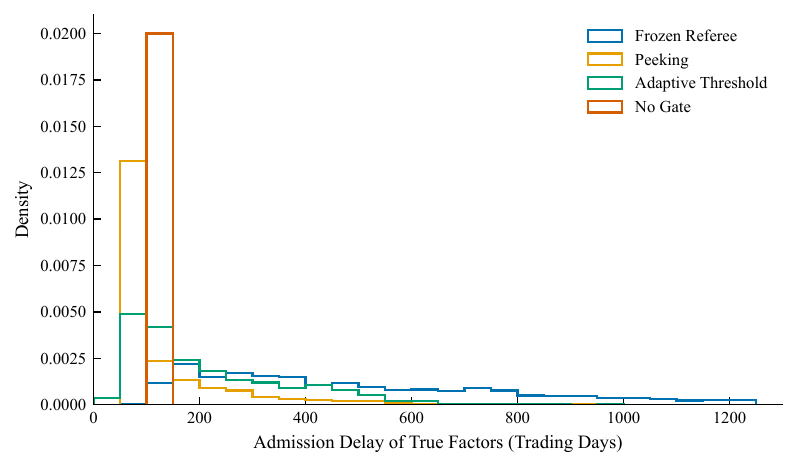}
\caption{Admission delay of realised-true factors by referee arm (LLM controller, four-family library, 2016 start, five seeds pooled). The no-gate spike is the pre-registered 126-day incubation.}
\label{fig:delay}
\end{figure}

This appendix presents detailed evidence for \S\ref{sec:results}. Tables
\ref{tbl:synthetic} and \ref{tbl:probe} present the synthetic factorial
and probe environment. Figure \ref{fig:false} plots Table
\ref{tbl:referee}. Tables \ref{tbl:anatomy} to \ref{tbl:retire} provide
zero-cost ledger analyses for the real-data sections. They cover the
realised edges of admitted factors, label-threshold sensitivity, and
after-admission and per-family labels. They also cover legitimate
comparators, counterfactual books, the period split, and retirement
outcomes. Table \ref{tbl:decomp} gives the Sharpe decomposition. Figure
\ref{fig:delay} shows the admission-delay distribution, and Table
\ref{tbl:regime} reports the regime timing. Tables \ref{tbl:paired} and
\ref{tbl:paired9} provide the paired controller contrasts behind Table
\ref{tbl:controller}.

\begin{table}[htbp]
\centering
\caption{Synthetic world with planted truth (four families, 1260 days,
five seeds). The table reports seed-mean realised yield and false
admissions per submission by referee arm and controller for the two LLM
families run there. It also reports the hidden-retry attacker's false
admissions.}\label{tbl:synthetic}
\begin{tabular}{@{}
  >{\raggedright\arraybackslash}p{(\linewidth - 8\tabcolsep) * \real{0.1739}}
  >{\raggedright\arraybackslash}p{(\linewidth - 8\tabcolsep) * \real{0.1739}}
  >{\centering\arraybackslash}p{(\linewidth - 8\tabcolsep) * \real{0.2174}}
  >{\centering\arraybackslash}p{(\linewidth - 8\tabcolsep) * \real{0.2174}}
  >{\centering\arraybackslash}p{(\linewidth - 8\tabcolsep) * \real{0.2174}}@{}}
\toprule\noalign{}
\begin{minipage}[b]{\linewidth}\raggedright
Family
\end{minipage} & \begin{minipage}[b]{\linewidth}\raggedright
Referee
\end{minipage} & \begin{minipage}[b]{\linewidth}\centering
Yield: Round-Robin / Bandit / LLM
\end{minipage} & \begin{minipage}[b]{\linewidth}\centering
False Admissions: Round-Robin / Bandit / LLM
\end{minipage} & \begin{minipage}[b]{\linewidth}\centering
Hidden Retries: False
\end{minipage} \\
\midrule\noalign{}
Gemini & Frozen & 0.095 / 0.096 / 0.096 & 0.000 / 0.000 / 0.000 &
0.000 \\
Gemini & Peeking & 0.060 / 0.065 / 0.063 & 0.314 / 0.300 / 0.315 &
0.341 \\
Gemini & Adaptive Threshold & 0.081 / 0.087 / 0.087 & 0.272 / 0.273 /
0.263 & 0.281 \\
Gemini & No Gate & 0.042 / 0.034 / 0.043 & 0.838 / 0.848 / 0.836 &
0.820 \\
GPT & Frozen & 0.095 / 0.096 / 0.096 & 0.000 / 0.000 / 0.000 & 0.000 \\
GPT & Peeking & 0.060 / 0.065 / 0.061 & 0.314 / 0.300 / 0.318 & 0.341 \\
GPT & Adaptive Threshold & 0.081 / 0.087 / 0.088 & 0.272 / 0.273 / 0.271
& 0.281 \\
GPT & No Gate & 0.042 / 0.034 / 0.041 & 0.838 / 0.848 / 0.839 & 0.820 \\
\bottomrule
\end{tabular}
\end{table}

\begin{table}[htbp]
\centering
\caption{Probe environment at the registered escalation price. For each
model family, the table reports mean intervention regret against the
oracle (edge units) \(\pm\) s.e. over 100 episodes for the fixed
diagnosis menu and authored probes. Paired differences are authored
minus menu within each episode. Holm correction covers the six evaluable
families.}\label{tbl:probe}
\begin{tabular}{@{}
  >{\raggedright\arraybackslash}p{(\linewidth - 8\tabcolsep) * \real{0.1667}}
  >{\centering\arraybackslash}p{(\linewidth - 8\tabcolsep) * \real{0.2083}}
  >{\centering\arraybackslash}p{(\linewidth - 8\tabcolsep) * \real{0.2083}}
  >{\centering\arraybackslash}p{(\linewidth - 8\tabcolsep) * \real{0.2083}}
  >{\centering\arraybackslash}p{(\linewidth - 8\tabcolsep) * \real{0.2083}}@{}}
\toprule\noalign{}
\begin{minipage}[b]{\linewidth}\raggedright
Family
\end{minipage} & \begin{minipage}[b]{\linewidth}\centering
Fixed Menu: Regret
\end{minipage} & \begin{minipage}[b]{\linewidth}\centering
Authored Probes: Regret
\end{minipage} & \begin{minipage}[b]{\linewidth}\centering
Paired Difference
\end{minipage} & \begin{minipage}[b]{\linewidth}\centering
Holm \(p\)
\end{minipage} \\
\midrule\noalign{}
DeepSeek & 0.616 \(\pm\) 0.098 & 0.228 \(\pm\) 0.064 &
\(-0.388\pm0.101\) & 0.001 \\
Muse & 0.651 \(\pm\) 0.084 & 0.403 \(\pm\) 0.064 & \(-0.248\pm0.070\) &
0.002 \\
Gemini & 0.692 \(\pm\) 0.098 & 0.458 \(\pm\) 0.117 & \(-0.233\pm0.085\)
& 0.025 \\
GPT & 0.526 \(\pm\) 0.078 & 0.459 \(\pm\) 0.097 & \(-0.067\pm0.072\) &
1.000 \\
Hunyuan & 0.663 \(\pm\) 0.097 & 0.642 \(\pm\) 0.109 & \(-0.021\pm0.069\)
& 1.000 \\
GLM & 0.585 \(\pm\) 0.093 & 0.659 \(\pm\) 0.148 & \(+0.074\pm0.114\) &
1.000 \\
Qwen & 0.415 \(\pm\) 0.103 & 0.415 \(\pm\) 0.103 & Inevaluable (No Probe
Produced) & \\
\bottomrule
\end{tabular}
\end{table}

\begin{table}[htbp]
\centering
\caption{Anatomy of admissions on the four-family library (means per
campaign over four starts \(\times\) five seeds). The table splits
ever-admitted factors by their realised post-submission mean rank-IC.
The first two classes together are realised false admissions. The last
column counts false admissions retired by the detector before campaign
end.}\label{tbl:anatomy}
\begin{tabular}{@{}
  >{\raggedright\arraybackslash}p{(\linewidth - 12\tabcolsep) * \real{0.2727}}
  >{\raggedleft\arraybackslash}p{(\linewidth - 12\tabcolsep) * \real{0.1212}}
  >{\raggedleft\arraybackslash}p{(\linewidth - 12\tabcolsep) * \real{0.1212}}
  >{\raggedleft\arraybackslash}p{(\linewidth - 12\tabcolsep) * \real{0.1212}}
  >{\raggedleft\arraybackslash}p{(\linewidth - 12\tabcolsep) * \real{0.1212}}
  >{\raggedleft\arraybackslash}p{(\linewidth - 12\tabcolsep) * \real{0.1212}}
  >{\raggedleft\arraybackslash}p{(\linewidth - 12\tabcolsep) * \real{0.1212}}@{}}
\toprule\noalign{}
\begin{minipage}[b]{\linewidth}\raggedright
Arm
\end{minipage} & \begin{minipage}[b]{\linewidth}\raggedleft
Admitted
\end{minipage} & \begin{minipage}[b]{\linewidth}\raggedleft
Edge \(<0\)
\end{minipage} & \begin{minipage}[b]{\linewidth}\raggedleft
\(0\le\) Edge \(<\delta\)
\end{minipage} & \begin{minipage}[b]{\linewidth}\raggedleft
Edge \(\ge\delta\)
\end{minipage} & \begin{minipage}[b]{\linewidth}\raggedleft
False Admissions
\end{minipage} & \begin{minipage}[b]{\linewidth}\raggedleft
Retired by the End
\end{minipage} \\
\midrule\noalign{}
Frozen, Round-Robin & 214.7 & 0.0 & 11.7 & 203.0 & 11.7 & 11.7 \\
Frozen, Bandit & 260.8 & 0.0 & 9.4 & 251.3 & 9.4 & 9.4 \\
Frozen, LLM (Gemini) & 257.0 & 0.0 & 11.7 & 245.3 & 11.7 & 11.7 \\
Frozen, LLM (DeepSeek) & 258.1 & 0.0 & 12.7 & 245.5 & 12.7 & 12.7 \\
Frozen, LLM (GPT) & 236.9 & 0.0 & 13.3 & 223.6 & 13.3 & 13.3 \\
Peeking, Round-Robin & 312.5 & 10.8 & 75.3 & 226.3 & 86.2 & 85.6 \\
Peeking, Bandit & 342.8 & 3.6 & 34.7 & 304.4 & 38.4 & 38.1 \\
Peeking, LLM (Gemini) & 354.9 & 1.2 & 36.1 & 317.6 & 37.4 & 37.0 \\
Adaptive Threshold, Round-Robin & 322.3 & 32.0 & 73.2 & 217.1 & 105.2 &
93.7 \\
Adaptive Threshold, Bandit & 327.6 & 11.6 & 39.4 & 276.6 & 51.0 &
47.7 \\
Adaptive Threshold, LLM (Gemini) & 330.6 & 6.0 & 37.5 & 287.1 & 43.5 &
42.8 \\
No Gate, Round-Robin & 398.0 & 90.8 & 105.2 & 202.0 & 196.0 & 150.8 \\
No Gate, Bandit & 398.0 & 92.0 & 97.8 & 208.2 & 189.8 & 148.1 \\
No Gate, LLM (Gemini) & 398.0 & 15.5 & 62.5 & 320.1 & 78.0 & 61.8 \\
No Gate, LLM (DeepSeek) & 398.0 & 49.5 & 77.0 & 271.4 & 126.6 & 97.5 \\
No Gate, LLM (GPT) & 398.0 & 74.2 & 94.9 & 228.9 & 169.1 & 126.6 \\
\bottomrule
\end{tabular}
\end{table}

\begin{table}[htbp]
\centering
\caption{Label-threshold sensitivity on the four-family library (means
over four starts \(\times\) five seeds). The table reports realised
yield per submission and realised false admissions per campaign at label
thresholds of \(0.010\), \(0.015\) (registered), and
\(0.020\).}\label{tbl:labelsens}
\begin{tabular}{@{}
  >{\raggedright\arraybackslash}p{(\linewidth - 6\tabcolsep) * \real{0.2222}}
  >{\raggedright\arraybackslash}p{(\linewidth - 6\tabcolsep) * \real{0.2222}}
  >{\centering\arraybackslash}p{(\linewidth - 6\tabcolsep) * \real{0.2778}}
  >{\centering\arraybackslash}p{(\linewidth - 6\tabcolsep) * \real{0.2778}}@{}}
\toprule\noalign{}
\begin{minipage}[b]{\linewidth}\raggedright
Referee
\end{minipage} & \begin{minipage}[b]{\linewidth}\raggedright
Controller
\end{minipage} & \begin{minipage}[b]{\linewidth}\centering
Yield at 0.010 / 0.015 / 0.020
\end{minipage} & \begin{minipage}[b]{\linewidth}\centering
False Admissions at 0.010 / 0.015 / 0.020
\end{minipage} \\
\midrule\noalign{}
Frozen & Round-Robin & 0.403 / 0.403 / 0.400 & 5.4 / 11.7 / 13.8 \\
Frozen & LLM (Gemini) & 0.471 / 0.471 / 0.466 & 5.7 / 11.7 / 14.7 \\
Peeking & Round-Robin & 0.407 / 0.406 / 0.402 & 78.5 / 86.2 / 90.2 \\
Peeking & LLM (Gemini) & 0.546 / 0.546 / 0.532 & 31.9 / 37.4 / 45.7 \\
Adaptive Threshold & Round-Robin & 0.390 / 0.390 / 0.387 & 97.1 / 105.2
/ 108.2 \\
Adaptive Threshold & LLM (Gemini) & 0.478 / 0.478 / 0.476 & 37.5 / 43.5
/ 47.0 \\
No Gate & Round-Robin & 0.370 / 0.367 / 0.361 & 185.8 / 196.0 / 200.5 \\
No Gate & LLM (Gemini) & 0.596 / 0.585 / 0.547 & 66.7 / 78.0 / 95.8 \\
\bottomrule
\end{tabular}
\end{table}

\begin{table}[htbp]
\centering
\caption{Four-family library results show realised false admissions
under three labels. Values are means per campaign over four starts
\(\times\) five seeds. \emph{Full} is the registered label. \emph{After}
judges each admitted factor only on days after admission. \(\delta_f\)
is each family's break-even (liquidity 0.012, volatility 0.017, momentum
0.031, reversal 0.074).}\label{tbl:postlabel}
\begin{tabular}{@{}
  >{\raggedright\arraybackslash}p{(\linewidth - 14\tabcolsep) * \real{0.1212}}
  >{\raggedright\arraybackslash}p{(\linewidth - 14\tabcolsep) * \real{0.1212}}
  >{\raggedleft\arraybackslash}p{(\linewidth - 14\tabcolsep) * \real{0.1212}}
  >{\raggedleft\arraybackslash}p{(\linewidth - 14\tabcolsep) * \real{0.1212}}
  >{\raggedleft\arraybackslash}p{(\linewidth - 14\tabcolsep) * \real{0.1212}}
  >{\raggedleft\arraybackslash}p{(\linewidth - 14\tabcolsep) * \real{0.1212}}
  >{\raggedleft\arraybackslash}p{(\linewidth - 14\tabcolsep) * \real{0.1212}}
  >{\centering\arraybackslash}p{(\linewidth - 14\tabcolsep) * \real{0.1515}}@{}}
\toprule\noalign{}
\begin{minipage}[b]{\linewidth}\raggedright
Referee
\end{minipage} & \begin{minipage}[b]{\linewidth}\raggedright
Controller
\end{minipage} & \begin{minipage}[b]{\linewidth}\raggedleft
Admitted
\end{minipage} & \begin{minipage}[b]{\linewidth}\raggedleft
False: Full
\end{minipage} & \begin{minipage}[b]{\linewidth}\raggedleft
False: After
\end{minipage} & \begin{minipage}[b]{\linewidth}\raggedleft
False at \(\delta_f\)
\end{minipage} & \begin{minipage}[b]{\linewidth}\raggedleft
Edge After
\end{minipage} & \begin{minipage}[b]{\linewidth}\centering
Yield: Full / After
\end{minipage} \\
\midrule\noalign{}
Frozen & Round-Robin & 214.7 & 11.7 & 17.6 & 96.0 & 0.032 & 0.403 /
0.380 \\
Frozen & Bandit & 260.8 & 9.4 & 14.4 & 114.3 & 0.032 & 0.474 / 0.460 \\
Frozen & LLM (Gemini) & 254.8 & 11.7 & 15.8 & 111.7 & 0.032 & 0.466 /
0.458 \\
Peeking & Round-Robin & 312.4 & 86.2 & 91.2 & 180.4 & 0.023 & 0.406 /
0.390 \\
Peeking & Bandit & 342.8 & 38.4 & 52.6 & 170.5 & 0.027 & 0.495 /
0.457 \\
Peeking & LLM (Gemini) & 354.6 & 37.4 & 41.5 & 152.9 & 0.029 & 0.545 /
0.511 \\
Adaptive Threshold & Round-Robin & 322.3 & 105.2 & 107.5 & 197.4 & 0.021
& 0.390 / 0.384 \\
Adaptive Threshold & Bandit & 327.6 & 51.0 & 54.4 & 185.8 & 0.027 &
0.442 / 0.430 \\
Adaptive Threshold & LLM (Gemini) & 330.2 & 43.5 & 44.4 & 158.2 & 0.029
& 0.478 / 0.466 \\
No Gate & Round-Robin & 398.0 & 196.0 & 195.6 & 276.2 & 0.016 & 0.367 /
0.358 \\
No Gate & Bandit & 398.0 & 189.8 & 194.1 & 265.6 & 0.015 & 0.372 /
0.352 \\
No Gate & LLM (Gemini) & 394.0 & 77.2 & 77.0 & 185.3 & 0.027 & 0.577 /
0.565 \\
\bottomrule
\end{tabular}
\end{table}

\begin{table}[htbp]
\centering
\caption{Nine-family library results show the same measures. They
exclude a per-family break-even because it is undefined for the five
fundamental families.}\label{tbl:postlabel9}
\begin{tabular}{@{}
  >{\raggedright\arraybackslash}p{(\linewidth - 12\tabcolsep) * \real{0.1379}}
  >{\raggedright\arraybackslash}p{(\linewidth - 12\tabcolsep) * \real{0.1379}}
  >{\raggedleft\arraybackslash}p{(\linewidth - 12\tabcolsep) * \real{0.1379}}
  >{\raggedleft\arraybackslash}p{(\linewidth - 12\tabcolsep) * \real{0.1379}}
  >{\raggedleft\arraybackslash}p{(\linewidth - 12\tabcolsep) * \real{0.1379}}
  >{\raggedleft\arraybackslash}p{(\linewidth - 12\tabcolsep) * \real{0.1379}}
  >{\centering\arraybackslash}p{(\linewidth - 12\tabcolsep) * \real{0.1724}}@{}}
\toprule\noalign{}
\begin{minipage}[b]{\linewidth}\raggedright
Referee
\end{minipage} & \begin{minipage}[b]{\linewidth}\raggedright
Controller
\end{minipage} & \begin{minipage}[b]{\linewidth}\raggedleft
Admitted
\end{minipage} & \begin{minipage}[b]{\linewidth}\raggedleft
False: Full
\end{minipage} & \begin{minipage}[b]{\linewidth}\raggedleft
False: After
\end{minipage} & \begin{minipage}[b]{\linewidth}\raggedleft
Edge After
\end{minipage} & \begin{minipage}[b]{\linewidth}\centering
Yield: Full / After
\end{minipage} \\
\midrule\noalign{}
Frozen & Round-Robin & 127.2 & 9.8 & 16.6 & 0.028 & 0.246 / 0.225 \\
Frozen & Bandit & 188.1 & 11.9 & 17.6 & 0.030 & 0.357 / 0.340 \\
Frozen & LLM (Gemini) & 198.6 & 14.6 & 20.3 & 0.030 & 0.377 / 0.364 \\
Peeking & Round-Robin & 256.9 & 127.3 & 131.8 & 0.016 & 0.244 / 0.233 \\
Peeking & Bandit & 295.8 & 97.0 & 107.2 & 0.020 & 0.360 / 0.330 \\
Adaptive Threshold & Round-Robin & 296.4 & 175.2 & 174.9 & 0.013 & 0.230
/ 0.228 \\
Adaptive Threshold & Bandit & 299.6 & 128.8 & 129.7 & 0.019 & 0.292 /
0.285 \\
No Gate & Round-Robin & 398.0 & 295.6 & 292.9 & 0.009 & 0.192 / 0.189 \\
No Gate & Bandit & 398.0 & 293.8 & 291.1 & 0.008 & 0.194 / 0.190 \\
\bottomrule
\end{tabular}
\end{table}

\begin{table}[htbp]
\centering
\caption{Four-family library results show fixed-horizon comparator
referees run as counterfactual programs on the frozen cell's recorded
submission stream. Values are means per campaign over four starts
\(\times\) five seeds. Each tests the post-submission mean once, 500 or
250 days after submission. Each applies Benjamini--Hochberg at 0.05
whenever a window closes (\emph{Daily}, deployable) or once at the
campaign's end (\emph{At End}, so no yield or
wait).}\label{tbl:comparator}
\begin{tabular}{@{}
  >{\raggedright\arraybackslash}p{(\linewidth - 16\tabcolsep) * \real{0.1373}}
  >{\raggedright\arraybackslash}p{(\linewidth - 16\tabcolsep) * \real{0.1961}}
  >{\raggedleft\arraybackslash}p{(\linewidth - 16\tabcolsep) * \real{0.1176}}
  >{\raggedleft\arraybackslash}p{(\linewidth - 16\tabcolsep) * \real{0.0980}}
  >{\raggedleft\arraybackslash}p{(\linewidth - 16\tabcolsep) * \real{0.0980}}
  >{\raggedleft\arraybackslash}p{(\linewidth - 16\tabcolsep) * \real{0.0980}}
  >{\raggedleft\arraybackslash}p{(\linewidth - 16\tabcolsep) * \real{0.0980}}
  >{\raggedleft\arraybackslash}p{(\linewidth - 16\tabcolsep) * \real{0.0882}}
  >{\raggedleft\arraybackslash}p{(\linewidth - 16\tabcolsep) * \real{0.0686}}@{}}
\toprule\noalign{}
\begin{minipage}[b]{\linewidth}\raggedright
Controller
\end{minipage} & \begin{minipage}[b]{\linewidth}\raggedright
Referee
\end{minipage} & \begin{minipage}[b]{\linewidth}\raggedleft
Admitted
\end{minipage} & \begin{minipage}[b]{\linewidth}\raggedleft
False: Full
\end{minipage} & \begin{minipage}[b]{\linewidth}\raggedleft
False: After
\end{minipage} & \begin{minipage}[b]{\linewidth}\raggedleft
Share False
\end{minipage} & \begin{minipage}[b]{\linewidth}\raggedleft
Edge After
\end{minipage} & \begin{minipage}[b]{\linewidth}\raggedleft
Yield
\end{minipage} & \begin{minipage}[b]{\linewidth}\raggedleft
Wait
\end{minipage} \\
\midrule\noalign{}
Round-Robin & Frozen (Recorded) & 214.7 & 11.7 & 17.6 & 0.08 & 0.032 &
0.403 & 439 \\
Round-Robin & Fixed 500 (Daily) & 208.0 & 16.4 & 22.2 & 0.11 & 0.029 &
0.358 & 500 \\
Round-Robin & Fixed 500 (At End) & 207.2 & 16.0 & 21.8 & 0.11 & 0.029 &
-- & -- \\
Round-Robin & Fixed 250 (Daily) & 187.4 & 16.6 & 19.4 & 0.10 & 0.030 &
0.314 & 250 \\
Bandit & Frozen (Recorded) & 260.8 & 9.4 & 14.4 & 0.06 & 0.032 & 0.474 &
408 \\
Bandit & Fixed 500 (Daily) & 262.6 & 12.2 & 20.4 & 0.08 & 0.028 & 0.441
& 500 \\
Bandit & Fixed 500 (At End) & 262.3 & 12.1 & 20.2 & 0.08 & 0.028 & -- &
-- \\
Bandit & Fixed 250 (Daily) & 239.6 & 9.8 & 13.3 & 0.06 & 0.031 & 0.404 &
250 \\
\bottomrule
\end{tabular}
\end{table}

\begin{table}[htbp]
\centering
\caption{Nine-family library results show the same
comparators.}\label{tbl:comparator9}
\begin{tabular}{@{}
  >{\raggedright\arraybackslash}p{(\linewidth - 16\tabcolsep) * \real{0.1373}}
  >{\raggedright\arraybackslash}p{(\linewidth - 16\tabcolsep) * \real{0.1961}}
  >{\raggedleft\arraybackslash}p{(\linewidth - 16\tabcolsep) * \real{0.1176}}
  >{\raggedleft\arraybackslash}p{(\linewidth - 16\tabcolsep) * \real{0.0980}}
  >{\raggedleft\arraybackslash}p{(\linewidth - 16\tabcolsep) * \real{0.0980}}
  >{\raggedleft\arraybackslash}p{(\linewidth - 16\tabcolsep) * \real{0.0980}}
  >{\raggedleft\arraybackslash}p{(\linewidth - 16\tabcolsep) * \real{0.0980}}
  >{\raggedleft\arraybackslash}p{(\linewidth - 16\tabcolsep) * \real{0.0882}}
  >{\raggedleft\arraybackslash}p{(\linewidth - 16\tabcolsep) * \real{0.0686}}@{}}
\toprule\noalign{}
\begin{minipage}[b]{\linewidth}\raggedright
Controller
\end{minipage} & \begin{minipage}[b]{\linewidth}\raggedright
Referee
\end{minipage} & \begin{minipage}[b]{\linewidth}\raggedleft
Admitted
\end{minipage} & \begin{minipage}[b]{\linewidth}\raggedleft
False: Full
\end{minipage} & \begin{minipage}[b]{\linewidth}\raggedleft
False: After
\end{minipage} & \begin{minipage}[b]{\linewidth}\raggedleft
Share False
\end{minipage} & \begin{minipage}[b]{\linewidth}\raggedleft
Edge After
\end{minipage} & \begin{minipage}[b]{\linewidth}\raggedleft
Yield
\end{minipage} & \begin{minipage}[b]{\linewidth}\raggedleft
Wait
\end{minipage} \\
\midrule\noalign{}
Round-Robin & Frozen (Recorded) & 127.2 & 9.8 & 16.6 & 0.12 & 0.028 &
0.246 & 543 \\
Round-Robin & Fixed 500 (Daily) & 150.2 & 41.4 & 49.7 & 0.33 & 0.021 &
0.214 & 500 \\
Round-Robin & Fixed 500 (At End) & 148.8 & 40.8 & 48.9 & 0.33 & 0.021 &
-- & -- \\
Round-Robin & Fixed 250 (Daily) & 115.0 & 30.2 & 33.8 & 0.29 & 0.023 &
0.159 & 250 \\
Bandit & Frozen (Recorded) & 188.1 & 11.9 & 17.6 & 0.09 & 0.030 & 0.357
& 436 \\
Bandit & Fixed 500 (Daily) & 207.8 & 33.2 & 44.0 & 0.22 & 0.024 & 0.330
& 500 \\
Bandit & Fixed 500 (At End) & 207.3 & 33.2 & 43.8 & 0.21 & 0.024 & -- &
-- \\
Bandit & Fixed 250 (Daily) & 177.2 & 25.3 & 29.2 & 0.17 & 0.027 & 0.287
& 250 \\
\bottomrule
\end{tabular}
\end{table}

\begin{table}[htbp]
\centering
\caption{Net Sharpe of counterfactual books on the round-robin frozen
cell's candidate sequence. Results use the four-family library and five
seeds per start. They report means with standard errors and differences
paired per seed. \emph{Retirement off} holds every factor admitted by
the frozen referee through the campaign's end. \emph{Fixed 500 (Daily)}
is Table \ref{tbl:comparator}`s deployable 500-day comparator. It is
held from its admission day until the runs' e-detector retires
it.}\label{tbl:cfbooks}
\begin{tabular}{@{}
  >{\raggedright\arraybackslash}p{(\linewidth - 12\tabcolsep) * \real{0.1176}}
  >{\centering\arraybackslash}p{(\linewidth - 12\tabcolsep) * \real{0.1471}}
  >{\centering\arraybackslash}p{(\linewidth - 12\tabcolsep) * \real{0.1471}}
  >{\centering\arraybackslash}p{(\linewidth - 12\tabcolsep) * \real{0.1471}}
  >{\centering\arraybackslash}p{(\linewidth - 12\tabcolsep) * \real{0.1471}}
  >{\centering\arraybackslash}p{(\linewidth - 12\tabcolsep) * \real{0.1471}}
  >{\centering\arraybackslash}p{(\linewidth - 12\tabcolsep) * \real{0.1471}}@{}}
\toprule\noalign{}
\begin{minipage}[b]{\linewidth}\raggedright
Start
\end{minipage} & \begin{minipage}[b]{\linewidth}\centering
No Gate
\end{minipage} & \begin{minipage}[b]{\linewidth}\centering
Frozen
\end{minipage} & \begin{minipage}[b]{\linewidth}\centering
Frozen, Retirement Off
\end{minipage} & \begin{minipage}[b]{\linewidth}\centering
Retirement Off Minus Frozen
\end{minipage} & \begin{minipage}[b]{\linewidth}\centering
Fixed 500 (Daily)
\end{minipage} & \begin{minipage}[b]{\linewidth}\centering
Fixed 500 Minus Frozen
\end{minipage} \\
\midrule\noalign{}
2016 & \(+0.74\pm0.02\) & \(+0.40\pm0.03\) & \(+0.41\pm0.03\) &
\(+0.00\pm0.00\) & \(+0.60\pm0.01\) & \(+0.20\pm0.03\) \\
2017 & \(+0.78\pm0.01\) & \(+0.46\pm0.01\) & \(+0.47\pm0.01\) &
\(+0.01\pm0.00\) & \(+0.59\pm0.02\) & \(+0.13\pm0.02\) \\
2018 & \(+0.75\pm0.01\) & \(+0.42\pm0.02\) & \(+0.42\pm0.02\) &
\(+0.01\pm0.00\) & \(+0.65\pm0.02\) & \(+0.24\pm0.02\) \\
2019 & \(+0.62\pm0.04\) & \(+0.49\pm0.05\) & \(+0.50\pm0.05\) &
\(+0.01\pm0.00\) & \(+0.61\pm0.02\) & \(+0.12\pm0.04\) \\
\bottomrule
\end{tabular}
\end{table}

\begin{table}[htbp]
\centering
\caption{Controller contrasts on the four-family library under the
frozen referee, split by candidates submitted before and from
2020-01-02. Each campaign has about 114 and 308 such candidates.
LLM-minus-baseline differences are paired by start and seed. Values are
means \(\pm\) s.e. over 20 pairs.}\label{tbl:period}
\begin{tabular}{@{}
  >{\raggedright\arraybackslash}p{(\linewidth - 10\tabcolsep) * \real{0.1087}}
  >{\raggedright\arraybackslash}p{(\linewidth - 10\tabcolsep) * \real{0.1957}}
  >{\centering\arraybackslash}p{(\linewidth - 10\tabcolsep) * \real{0.1739}}
  >{\centering\arraybackslash}p{(\linewidth - 10\tabcolsep) * \real{0.1739}}
  >{\centering\arraybackslash}p{(\linewidth - 10\tabcolsep) * \real{0.1739}}
  >{\centering\arraybackslash}p{(\linewidth - 10\tabcolsep) * \real{0.1739}}@{}}
\toprule\noalign{}
\begin{minipage}[b]{\linewidth}\raggedright
Family
\end{minipage} & \begin{minipage}[b]{\linewidth}\raggedright
Contrast
\end{minipage} & \begin{minipage}[b]{\linewidth}\centering
Yield Diff. Pre-2020
\end{minipage} & \begin{minipage}[b]{\linewidth}\centering
Yield Diff. Post-2020
\end{minipage} & \begin{minipage}[b]{\linewidth}\centering
False Adm. Diff. Pre-2020
\end{minipage} & \begin{minipage}[b]{\linewidth}\centering
False Adm. Diff. Post-2020
\end{minipage} \\
\midrule\noalign{}
Gemini & LLM \(-\) Round-Robin & \(+0.009\pm0.005\) & \(+0.088\pm0.004\)
& \(-0.2\pm0.1\) & \(+0.1\pm0.2\) \\
Gemini & LLM \(-\) Bandit & \(+0.026\pm0.006\) & \(-0.016\pm0.004\) &
\(+0.3\pm0.2\) & \(+1.9\pm0.7\) \\
DeepSeek & LLM \(-\) Round-Robin & \(+0.028\pm0.009\) &
\(+0.097\pm0.004\) & \(-0.1\pm0.1\) & \(+1.1\pm0.6\) \\
DeepSeek & LLM \(-\) Bandit & \(+0.045\pm0.011\) & \(-0.007\pm0.004\) &
\(+0.5\pm0.2\) & \(+2.8\pm0.9\) \\
GPT & LLM \(-\) Round-Robin & \(+0.008\pm0.003\) & \(+0.050\pm0.004\) &
\(-0.1\pm0.1\) & \(+1.7\pm0.4\) \\
GPT & LLM \(-\) Bandit & \(+0.025\pm0.006\) & \(-0.054\pm0.003\) &
\(+0.5\pm0.2\) & \(+3.4\pm0.8\) \\
\bottomrule
\end{tabular}
\end{table}

\begin{table}[htbp]
\centering
\caption{Retirement outcomes on real data using the four-family library,
frozen referee, and round-robin controller. Results cover four starts
\(\times\) five seeds. They report factors retired per campaign and
family. For retired factors, results also report mean daily rank-IC
during the held window and the remaining campaign. These factors had at
least 21 days held and 21 days left in the
campaign.}\label{tbl:retire}
\begin{tabular}{@{}
  >{\raggedright\arraybackslash}p{(\linewidth - 14\tabcolsep) * \real{0.1765}}
  >{\raggedleft\arraybackslash}p{(\linewidth - 14\tabcolsep) * \real{0.1176}}
  >{\raggedleft\arraybackslash}p{(\linewidth - 14\tabcolsep) * \real{0.1176}}
  >{\raggedleft\arraybackslash}p{(\linewidth - 14\tabcolsep) * \real{0.1176}}
  >{\raggedleft\arraybackslash}p{(\linewidth - 14\tabcolsep) * \real{0.1176}}
  >{\raggedleft\arraybackslash}p{(\linewidth - 14\tabcolsep) * \real{0.1176}}
  >{\raggedleft\arraybackslash}p{(\linewidth - 14\tabcolsep) * \real{0.1176}}
  >{\raggedleft\arraybackslash}p{(\linewidth - 14\tabcolsep) * \real{0.1176}}@{}}
\toprule\noalign{}
\begin{minipage}[b]{\linewidth}\raggedright
Family
\end{minipage} & \begin{minipage}[b]{\linewidth}\raggedleft
Retired
\end{minipage} & \begin{minipage}[b]{\linewidth}\raggedleft
Share Retired
\end{minipage} & \begin{minipage}[b]{\linewidth}\raggedleft
Days Held
\end{minipage} & \begin{minipage}[b]{\linewidth}\raggedleft
IC While Held
\end{minipage} & \begin{minipage}[b]{\linewidth}\raggedleft
IC After
\end{minipage} & \begin{minipage}[b]{\linewidth}\raggedleft
After \(<\delta\)
\end{minipage} & \begin{minipage}[b]{\linewidth}\raggedleft
After \(<0\)
\end{minipage} \\
\midrule\noalign{}
Liquidity & 12.1 & 0.24 & 207 & \(-0.0053\) & \(+0.0040\) & 1.00 &
0.05 \\
Momentum & 0.5 & 1.00 & 354 & \(-0.0131\) & \(+0.0035\) & 0.89 & 0.22 \\
Reversal & 31.8 & 0.36 & 838 & \(+0.0145\) & \(+0.0338\) & 0.00 &
0.00 \\
All & 44.3 & 0.21 & 661 & \(+0.0088\) & \(+0.0254\) & 0.28 & 0.01 \\
\bottomrule
\end{tabular}
\end{table}

\begin{table}[htbp]
\centering
\caption{Net Sharpe of counterfactual books under the canonical
construction. The books are frozen, ungated, (a) the referee's factors
on the ungated schedule, and (b) a fixed random subset of ungated
factors matching the referee's count. \emph{Breadth} is no gate less
(b). \emph{Selection} is (b) less (a). \emph{Earliness} is (a) less
frozen. The three sum to the gap. Sharpe here uses 252
days.}\label{tbl:decomp}
\begin{tabular}{@{}
  >{\raggedright\arraybackslash}p{(\linewidth - 14\tabcolsep) * \real{0.1250}}
  >{\raggedright\arraybackslash}p{(\linewidth - 14\tabcolsep) * \real{0.1250}}
  >{\raggedright\arraybackslash}p{(\linewidth - 14\tabcolsep) * \real{0.1250}}
  >{\raggedright\arraybackslash}p{(\linewidth - 14\tabcolsep) * \real{0.1250}}
  >{\raggedright\arraybackslash}p{(\linewidth - 14\tabcolsep) * \real{0.1250}}
  >{\raggedright\arraybackslash}p{(\linewidth - 14\tabcolsep) * \real{0.1250}}
  >{\raggedright\arraybackslash}p{(\linewidth - 14\tabcolsep) * \real{0.1250}}
  >{\raggedright\arraybackslash}p{(\linewidth - 14\tabcolsep) * \real{0.1250}}@{}}
\toprule\noalign{}
\begin{minipage}[b]{\linewidth}\raggedright
Start
\end{minipage} & \begin{minipage}[b]{\linewidth}\raggedright
Frozen
\end{minipage} & \begin{minipage}[b]{\linewidth}\raggedright
No Gate
\end{minipage} & \begin{minipage}[b]{\linewidth}\raggedright
(a)
\end{minipage} & \begin{minipage}[b]{\linewidth}\raggedright
(b)
\end{minipage} & \begin{minipage}[b]{\linewidth}\raggedright
Breadth
\end{minipage} & \begin{minipage}[b]{\linewidth}\raggedright
Selection
\end{minipage} & \begin{minipage}[b]{\linewidth}\raggedright
Earliness
\end{minipage} \\
\midrule\noalign{}
2016 & \(+0.40\) & \(+0.74\) & \(+0.63\) & \(+0.72\) & \(+0.02\) &
\(+0.09\) & \(+0.23\) \\
2017 & \(+0.46\) & \(+0.78\) & \(+0.65\) & \(+0.75\) & \(+0.04\) &
\(+0.10\) & \(+0.19\) \\
2018 & \(+0.42\) & \(+0.75\) & \(+0.67\) & \(+0.71\) & \(+0.04\) &
\(+0.04\) & \(+0.25\) \\
2019 & \(+0.49\) & \(+0.62\) & \(+0.53\) & \(+0.57\) & \(+0.05\) &
\(+0.04\) & \(+0.04\) \\
\bottomrule
\end{tabular}
\end{table}

\begin{table}[htbp]
\centering
\caption{Nine-family library with the round-robin controller, 2016
start, and five pooled seeds. The table reports sleeves ever held by the
book. It also reports each family's mean first-held campaign day, of
2,589. Finally, it reports the share of holding windows with midpoints
after 2020-01-02, campaign day 975.}\label{tbl:regime}
\begin{tabular}{@{}
  >{\raggedright\arraybackslash}p{(\linewidth - 12\tabcolsep) * \real{0.2203}}
  >{\raggedleft\arraybackslash}p{(\linewidth - 12\tabcolsep) * \real{0.1186}}
  >{\raggedleft\arraybackslash}p{(\linewidth - 12\tabcolsep) * \real{0.1525}}
  >{\raggedleft\arraybackslash}p{(\linewidth - 12\tabcolsep) * \real{0.1186}}
  >{\raggedleft\arraybackslash}p{(\linewidth - 12\tabcolsep) * \real{0.1186}}
  >{\raggedleft\arraybackslash}p{(\linewidth - 12\tabcolsep) * \real{0.1525}}
  >{\raggedleft\arraybackslash}p{(\linewidth - 12\tabcolsep) * \real{0.1186}}@{}}
\toprule\noalign{}
\begin{minipage}[b]{\linewidth}\raggedright
Family
\end{minipage} & \begin{minipage}[b]{\linewidth}\raggedleft
Certified: Sleeves
\end{minipage} & \begin{minipage}[b]{\linewidth}\raggedleft
First Held (Mean Day)
\end{minipage} & \begin{minipage}[b]{\linewidth}\raggedleft
After 2020
\end{minipage} & \begin{minipage}[b]{\linewidth}\raggedleft
Ungated: Sleeves
\end{minipage} & \begin{minipage}[b]{\linewidth}\raggedleft
First Held (Mean Day)
\end{minipage} & \begin{minipage}[b]{\linewidth}\raggedleft
After 2020
\end{minipage} \\
\midrule\noalign{}
Growth & 42 & 1151 & 100\% & 235 & 1365 & 83\% \\
Investment & 0 & & & 412 & 1419 & 76\% \\
Liquidity & 126 & 1511 & 100\% & 213 & 1264 & 83\% \\
Momentum & 0 & & & 354 & 1561 & 84\% \\
Quality & 45 & 1177 & 100\% & 306 & 1541 & 86\% \\
Reversal & 237 & 1722 & 100\% & 164 & 1266 & 100\% \\
Size & 2 & 1842 & 100\% & 321 & 1242 & 65\% \\
Value & 171 & 1881 & 99\% & 184 & 1151 & 68\% \\
Volatility & 203 & 1509 & 100\% & 151 & 1190 & 100\% \\
All Families & 826 & 1612 & 100\% & 2340 & 1366 & 81\% \\
\bottomrule
\end{tabular}
\end{table}

\begin{table}[htbp]
\centering
\caption{Paired realised-yield contrasts under the frozen referee on the
four-family library. Contrasts compare the LLM controller with
round-robin and the bandit. They report the mean difference across five
common-random-number seeds and its standard error. They also report the
Holm-adjusted two-sided \(p\) within the model family, over 24
tests.}\label{tbl:paired}
\begin{tabular}{@{}
  >{\raggedright\arraybackslash}p{(\linewidth - 10\tabcolsep) * \real{0.1667}}
  >{\raggedleft\arraybackslash}p{(\linewidth - 10\tabcolsep) * \real{0.1000}}
  >{\raggedleft\arraybackslash}p{(\linewidth - 10\tabcolsep) * \real{0.2500}}
  >{\raggedleft\arraybackslash}p{(\linewidth - 10\tabcolsep) * \real{0.1167}}
  >{\raggedleft\arraybackslash}p{(\linewidth - 10\tabcolsep) * \real{0.2500}}
  >{\raggedleft\arraybackslash}p{(\linewidth - 10\tabcolsep) * \real{0.1167}}@{}}
\toprule\noalign{}
\begin{minipage}[b]{\linewidth}\raggedright
Family
\end{minipage} & \begin{minipage}[b]{\linewidth}\raggedleft
Start
\end{minipage} & \begin{minipage}[b]{\linewidth}\raggedleft
LLM \(-\) Round-Robin Yield
\end{minipage} & \begin{minipage}[b]{\linewidth}\raggedleft
Holm \(p\)
\end{minipage} & \begin{minipage}[b]{\linewidth}\raggedleft
LLM \(-\) Bandit Yield
\end{minipage} & \begin{minipage}[b]{\linewidth}\raggedleft
Holm \(p\)
\end{minipage} \\
\midrule\noalign{}
DeepSeek & 2016 & \(+0.095\pm0.007\) & 0.006 & \(+0.039\pm0.007\) &
0.124 \\
DeepSeek & 2017 & \(+0.079\pm0.005\) & 0.005 & \(+0.007\pm0.004\) &
1.000 \\
DeepSeek & 2018 & \(+0.080\pm0.010\) & 0.034 & \(-0.004\pm0.006\) &
1.000 \\
DeepSeek & 2019 & \(+0.070\pm0.006\) & 0.012 & \(-0.002\pm0.007\) &
1.000 \\
Gemini & 2016 & \(+0.064\pm0.009\) & 0.122 & \(+0.008\pm0.007\) &
1.000 \\
Gemini & 2017 & \(+0.070\pm0.005\) & 0.008 & \(-0.002\pm0.002\) &
1.000 \\
Gemini & 2018 & \(+0.076\pm0.003\) & 0.001 & \(-0.009\pm0.003\) &
1.000 \\
Gemini & 2019 & \(+0.063\pm0.006\) & 0.033 & \(-0.009\pm0.012\) &
1.000 \\
GPT & 2016 & \(+0.046\pm0.005\) & 0.039 & \(-0.010\pm0.005\) & 1.000 \\
GPT & 2017 & \(+0.038\pm0.004\) & 0.029 & \(-0.034\pm0.006\) & 0.156 \\
GPT & 2018 & \(+0.045\pm0.008\) & 0.136 & \(-0.039\pm0.005\) & 0.060 \\
GPT & 2019 & \(+0.028\pm0.008\) & 0.537 & \(-0.043\pm0.007\) & 0.111 \\
\bottomrule
\end{tabular}
\end{table}

\begin{table}[htbp]
\centering
\caption{The same contrasts on the nine-family
library.}\label{tbl:paired9}
\begin{tabular}{@{}
  >{\raggedright\arraybackslash}p{(\linewidth - 10\tabcolsep) * \real{0.1667}}
  >{\raggedleft\arraybackslash}p{(\linewidth - 10\tabcolsep) * \real{0.1000}}
  >{\raggedleft\arraybackslash}p{(\linewidth - 10\tabcolsep) * \real{0.2500}}
  >{\raggedleft\arraybackslash}p{(\linewidth - 10\tabcolsep) * \real{0.1167}}
  >{\raggedleft\arraybackslash}p{(\linewidth - 10\tabcolsep) * \real{0.2500}}
  >{\raggedleft\arraybackslash}p{(\linewidth - 10\tabcolsep) * \real{0.1167}}@{}}
\toprule\noalign{}
\begin{minipage}[b]{\linewidth}\raggedright
Family
\end{minipage} & \begin{minipage}[b]{\linewidth}\raggedleft
Start
\end{minipage} & \begin{minipage}[b]{\linewidth}\raggedleft
LLM \(-\) Round-Robin Yield
\end{minipage} & \begin{minipage}[b]{\linewidth}\raggedleft
Holm \(p\)
\end{minipage} & \begin{minipage}[b]{\linewidth}\raggedleft
LLM \(-\) Bandit Yield
\end{minipage} & \begin{minipage}[b]{\linewidth}\raggedleft
Holm \(p\)
\end{minipage} \\
\midrule\noalign{}
DeepSeek & 2016 & \(+0.142\pm0.004\) & 0.000 & \(+0.029\pm0.007\) &
0.178 \\
DeepSeek & 2017 & \(+0.120\pm0.009\) & 0.003 & \(+0.011\pm0.008\) &
1.000 \\
DeepSeek & 2018 & \(+0.148\pm0.007\) & 0.001 & \(+0.026\pm0.004\) &
0.072 \\
DeepSeek & 2019 & \(+0.090\pm0.009\) & 0.009 & \(-0.010\pm0.005\) &
1.000 \\
Gemini & 2016 & \(+0.163\pm0.003\) & 0.000 & \(+0.050\pm0.005\) &
0.012 \\
Gemini & 2017 & \(+0.134\pm0.007\) & 0.001 & \(+0.026\pm0.006\) &
0.260 \\
Gemini & 2018 & \(+0.159\pm0.013\) & 0.005 & \(+0.036\pm0.010\) &
0.310 \\
Gemini & 2019 & \(+0.092\pm0.015\) & 0.068 & \(-0.009\pm0.012\) &
1.000 \\
GPT & 2016 & \(+0.059\pm0.010\) & 0.066 & \(-0.054\pm0.007\) & 0.032 \\
GPT & 2017 & \(+0.052\pm0.003\) & 0.001 & \(-0.057\pm0.009\) & 0.064 \\
GPT & 2018 & \(+0.051\pm0.010\) & 0.116 & \(-0.072\pm0.008\) & 0.021 \\
GPT & 2019 & \(+0.018\pm0.010\) & 1.000 & \(-0.082\pm0.006\) & 0.004 \\
\bottomrule
\end{tabular}
\end{table}

\section{Environment, amendments and information
parity}\label{app:amendments}

The 540 cells comprise 320 baseline cells and 220 LLM cells (Table
\ref{tbl:coverage}). Table \ref{tbl:amendments} lists every dated
amendment to the pre-registered design.

\begin{table}[htbp]
\centering
\caption{Coverage of the real-data factorial. Every row covers four
start years and five seeds. The last row is the confirmatory extension
run after the factorial.}\label{tbl:coverage}
\begin{tabular}{@{}
  >{\raggedright\arraybackslash}p{(\linewidth - 10\tabcolsep) * \real{0.1600}}
  >{\raggedright\arraybackslash}p{(\linewidth - 10\tabcolsep) * \real{0.1600}}
  >{\raggedright\arraybackslash}p{(\linewidth - 10\tabcolsep) * \real{0.1600}}
  >{\raggedright\arraybackslash}p{(\linewidth - 10\tabcolsep) * \real{0.1600}}
  >{\centering\arraybackslash}p{(\linewidth - 10\tabcolsep) * \real{0.2000}}
  >{\raggedleft\arraybackslash}p{(\linewidth - 10\tabcolsep) * \real{0.1600}}@{}}
\toprule\noalign{}
\begin{minipage}[b]{\linewidth}\raggedright
Controllers
\end{minipage} & \begin{minipage}[b]{\linewidth}\raggedright
Model Families
\end{minipage} & \begin{minipage}[b]{\linewidth}\raggedright
Referee Arms
\end{minipage} & \begin{minipage}[b]{\linewidth}\raggedright
Libraries
\end{minipage} & \begin{minipage}[b]{\linewidth}\centering
Starts \(\times\) Seeds
\end{minipage} & \begin{minipage}[b]{\linewidth}\raggedleft
Cells
\end{minipage} \\
\midrule\noalign{}
Round-Robin, Bandit & None & All Four & Both & \(4\times5\) & 320 \\
LLM & Gemini, DeepSeek, GPT & Frozen & Both & \(4\times5\) & 120 \\
LLM & Gemini & Peeking, Adaptive Threshold, No Gate & Four-Family &
\(4\times5\) & 60 \\
LLM & DeepSeek, GPT & No Gate & Four-Family & \(4\times5\) & 40 \\
Total & & & & & 540 \\
\bottomrule
\end{tabular}
\end{table}

\emph{The probe environment.} The five held-out fault types include
collapsed cross-sectional dispersion, possible crowding or regime-change
decay, and lagged restatement of fundamentals. They also include
sibling-factor contamination and a cost-channel fault. The fixed menu
contains three instruments: a data-health canary, a crowding proxy and
an IC changepoint test. The controller chooses one of five
interventions: retire the factor, deprecate or pause its family, repair
the data pipeline, or pay to escalate. A transparent economic model
gives each intervention's after-cost value in each true state over 126
days. Regret is the oracle's value less the chosen intervention's value.

\begin{table}[htbp]
\centering
\caption{Dated amendments to the pre-registered design. A5 and A9 were
never issued. Every book amendment used zero-cost re-booking of the
recorded cells (Proposition 2).}\label{tbl:amendments}
\begin{tabular}{@{}
  >{\raggedright\arraybackslash}p{(\linewidth - 6\tabcolsep) * \real{0.0882}}
  >{\raggedright\arraybackslash}p{(\linewidth - 6\tabcolsep) * \real{0.1765}}
  >{\raggedright\arraybackslash}p{(\linewidth - 6\tabcolsep) * \real{0.5588}}
  >{\raggedright\arraybackslash}p{(\linewidth - 6\tabcolsep) * \real{0.1765}}@{}}
\toprule\noalign{}
\begin{minipage}[b]{\linewidth}\raggedright
Id
\end{minipage} & \begin{minipage}[b]{\linewidth}\raggedright
Date
\end{minipage} & \begin{minipage}[b]{\linewidth}\raggedright
What Changed
\end{minipage} & \begin{minipage}[b]{\linewidth}\raggedright
Affects
\end{minipage} \\
\midrule\noalign{}
A1 & 2026-09-02 & LLM costs metered at billed prices; spend cap \$335 &
No (accounting) \\
A2 & 2026-09-02 & Deterministic iteration order; longer retry window &
No (engineering) \\
A3 & 2026-09-03 & Escalation-price sensitivity computed by re-scoring
recorded episodes & No (analysis) \\
A4 & 2026-09-04 & Abstract effort pins realised through the measured
vendor map & No (runtime) \\
A6 & 2026-09-04 & Probe environment audited; two world defects fixed
(world v2) & Probe Environment \\
A7 & 2026-09-05 & Real-data world specification on the CSI 500 &
Real-Data World \\
A8 & 2026-09-07 & Nine-family library; trading-frequency slice &
Library \\
A10 & 2026-09-07 & Framework consolidation before the restart; earlier
recorded runs not used & All Real-Data Runs \\
A10b & 2026-09-07 & Ex-ante book controls (turnover cap, index hedge);
canonical constructed book & Book \\
A11 & 2026-09-11 & Certification, persistence and execution separated;
per-sleeve calibrated holding period & Book \\
A11b & 2026-09-11 & Five-frequency calendar grid; dynamic re-calibration
with a cool-down & Book \\
A11c & 2026-09-13 & Conversion rate fixed as a market constant after a
ledger audit of the first pass & Book \\
\bottomrule
\end{tabular}
\end{table}

\begin{table}[htbp]
\centering
\caption{The controller ladder. Every rung uses the same library,
submission budget and referee. The rungs differ in policy and, at the
top rung, inputs and actions. Thus, the contrasts compare controller
packages.}\label{tbl:ladder}
\begin{tabular}{@{}
  >{\raggedright\arraybackslash}p{(\linewidth - 8\tabcolsep) * \real{0.1333}}
  >{\raggedright\arraybackslash}p{(\linewidth - 8\tabcolsep) * \real{0.2333}}
  >{\raggedright\arraybackslash}p{(\linewidth - 8\tabcolsep) * \real{0.2556}}
  >{\raggedright\arraybackslash}p{(\linewidth - 8\tabcolsep) * \real{0.1667}}
  >{\raggedright\arraybackslash}p{(\linewidth - 8\tabcolsep) * \real{0.2111}}@{}}
\toprule\noalign{}
\begin{minipage}[b]{\linewidth}\raggedright
Rung
\end{minipage} & \begin{minipage}[b]{\linewidth}\raggedright
Allocation Across Families
\end{minipage} & \begin{minipage}[b]{\linewidth}\raggedright
Response to an Alarm
\end{minipage} & \begin{minipage}[b]{\linewidth}\raggedright
Instruments
\end{minipage} & \begin{minipage}[b]{\linewidth}\raggedright
Inputs Beyond Family Statistics
\end{minipage} \\
\midrule\noalign{}
Round-robin (C0) & Cycles through families & Scripted: retire and refill
the same family & Fixed menu & None \\
Discounted-UCB bandit (C1) & Reward per family: admissions less
retirements, discounted & Scripted, as C0 & Fixed menu & None \\
LLM controller (C2) & Reads statistics, family profiles and memory &
Diagnosis (idiosyncratic death, crowding, regime shift, data breakage)
or escalation; chooses the response & Authors, tests and mounts probes &
Family profiles, memory, the reasons for retirements \\
\bottomrule
\end{tabular}
\end{table}

\emph{Information parity.} All controllers receive the same numeric
family statistics. The LLM controller also receives family profiles,
open candidates, paused families and memory. The controller contrasts
therefore compare packages. No controller sees the book. The text names
model versions only by family, while the run manifests pin them.

\emph{The break-even figure in the family profiles.} The family profiles
contain a scaling error that does not change how the controller
contrasts are read. Their break-even ICs used the regression slope of
\S\ref{sec:kappa}, not the average conversion. Thus, every figure is
between a quarter and a third of the corrected value. The error is a
common scale. The profiles rank families correctly and understate only
the level. The referee never reads the profile. The controller uses it
only to weigh families against one another.

\section{Data, point-in-time rules and the factor
library}\label{app:data}

Every quantity used on a day is known before that day's position is
taken. Membership uses the latest index-weight snapshot dated at or
before the previous trading day. Prices are adjusted by the vendor's
cumulative factor. Halts are missing values. A name that closed at its
price limit (\(|\Delta p|\ge9.5\%\)) on day \(t-1\) is excluded from the
day-\(t\) universe. Fundamentals become visible on their announcement
date and have a 400-day staleness cut. The panel covers 2834 trading
days. It includes 1356 stocks that were ever members. Every
vendor-extract partition has a SHA-256 in a provenance manifest.

\emph{The factor library.} The library is fixed in advance, keeping the
hypothesis universe finite as Proposition 1 requires. Each family
represents an economic hypothesis with a fixed sign and parameter grid.
Each submission draws one specification from that grid (Table
\ref{tbl:library}).

\begin{table}[htbp]
\centering
\caption{The nine factor families (the four-family library is the first
four). Prices use adjusted closes. Fundamentals are
announcement-aligned. The grid is enumerable, fixing the hypothesis
universe before the campaign.}\label{tbl:library}
\begin{tabular}{@{}
  >{\raggedright\arraybackslash}p{(\linewidth - 6\tabcolsep) * \real{0.1250}}
  >{\raggedright\arraybackslash}p{(\linewidth - 6\tabcolsep) * \real{0.4896}}
  >{\raggedright\arraybackslash}p{(\linewidth - 6\tabcolsep) * \real{0.3021}}
  >{\raggedleft\arraybackslash}p{(\linewidth - 6\tabcolsep) * \real{0.0833}}@{}}
\toprule\noalign{}
\begin{minipage}[b]{\linewidth}\raggedright
Family
\end{minipage} & \begin{minipage}[b]{\linewidth}\raggedright
Signal (sign)
\end{minipage} & \begin{minipage}[b]{\linewidth}\raggedright
Parameters Drawn
\end{minipage} & \begin{minipage}[b]{\linewidth}\raggedleft
Specs
\end{minipage} \\
\midrule\noalign{}
Momentum & \(\log(P_{t-1-s}/P_{t-1-s-L})\) (\(+\)) & \(L\in[40,250]\),
skip \(s\in[0,21]\) & 4,642 \\
Short-Horizon Reversal & \(-\log(P_{t-1}/P_{t-1-L})\) (\(+\) after sign)
& \(L\in[3,30]\); turnover-scaled or not & 56 \\
Volatility & \(-\)(std, downside std or max) of daily returns over \(L\)
(\(+\) after sign) & \(L\in[10,130]\); three kinds & 363 \\
Liquidity & \(-\)mean turnover over \(L\), or Amihud illiquidity (\(+\)
after sign) & \(L\in[10,130]\); two kinds & 242 \\
Value & Book, earnings, sales or dividends to price, smoothed over \(L\)
(\(+\)) & \(L\in[1,21]\); four kinds & 84 \\
Size & \(-\log\) market capitalisation, circulating or total (\(+\)
after sign) & \(L\in[1,21]\); two kinds & 42 \\
Quality & ROE, quarterly ROE, ROA, gross or net margin (\(+\)) &
\(L\in[1,63]\); five kinds & 315 \\
Growth & Net profit, revenue or EPS growth, year on year (\(+\)) &
\(L\in[1,63]\); three kinds & 189 \\
Investment (asset Growth) & Asset or equity growth, year on year (\(-\))
& \(L\in[1,63]\); two kinds & 126 \\
\bottomrule
\end{tabular}
\end{table}

\clearpage
\end{appendices}


\begin{CSLReferences}{1}{1}
\bibitem[\citeproctext]{ref-ahmadi2021fallacy}
Ahmadi-Javid, Amir, and Mohsen Ebadi. 2021. {``Economic Design of
Memory-Type Control Charts: The Fallacy of the Formula Proposed by
{L}orenzen and {V}ance (1986).''} \emph{Computational Statistics} 36
(1): 661--90. \url{https://doi.org/10.1007/s00180-020-01019-6}.

\bibitem[\citeproctext]{ref-bailey2014dsr}
Bailey, David H., and Marcos López de Prado. 2014. {``The Deflated
Sharpe Ratio: Correcting for Selection Bias, Backtest Overfitting, and
Non-Normality.''} \emph{The Journal of Portfolio Management} 40 (5):
94--107. \url{https://doi.org/10.3905/jpm.2014.40.5.094}.

\bibitem[\citeproctext]{ref-chen2026rsi}
Chen, Mingguang, Licheng Wang, and Bo Qu. 2026. \emph{Recursive
Self-Improvement in {AI}: From Bounded Self-Refinement to Autonomous
Research Loops}. \url{https://arxiv.org/abs/2607.07663}.

\bibitem[\citeproctext]{ref-clerico2026}
Clerico, Eugenio. 2026. \emph{Sequential Testing of Conditionally
Constrained Hypotheses}. \url{https://arxiv.org/abs/2606.06769}.

\bibitem[\citeproctext]{ref-dandapanthula2025eop}
Dandapanthula, Sanjit, and Aaditya Ramdas. 2025. \emph{Multiple Testing
in Multi-Stream Sequential Change Detection}.
\url{https://arxiv.org/abs/2501.04130}.

\bibitem[\citeproctext]{ref-fischer2024online}
Fischer, Lasse, Ziyu Xu, and Aaditya Ramdas. 2024. \emph{An Online
Generalization of the (e-){B}enjamini-{H}ochberg Procedure}.
\url{https://arxiv.org/abs/2407.20683}.

\bibitem[\citeproctext]{ref-replaygap2026}
Gonuguntla, Ashritha. 2026. \emph{The Replay Gap: Static Evaluation of
Model Switching in {LLM} Agents Scores the Wrong World}.
\url{https://arxiv.org/abs/2608.08239}.

\bibitem[\citeproctext]{ref-harvey2016}
Harvey, Campbell R., Yan Liu, and Heqing Zhu. 2016. {``\ldots And the
Cross-Section of Expected Returns.''} \emph{The Review of Financial
Studies} 29 (1): 5--68. \url{https://doi.org/10.1093/rfs/hhv059}.

\bibitem[\citeproctext]{ref-deheide2026}
Heide, Rianne de. 2026. \emph{Dynamic \(e\)-Closure for Online
Hypotheses with Any-Time-Valid Evidence: Closure Principles and
Projective Mergers}. \url{https://arxiv.org/abs/2608.09927}.

\bibitem[\citeproctext]{ref-rdagent2025}
Li, Yuante, Xu Yang, Xiao Yang, et al. 2025. \emph{{R\&D-Agent-Quant}: A
Multi-Agent Framework for Data-Centric Factors and Model Joint
Optimization}. \url{https://arxiv.org/abs/2505.15155}.

\bibitem[\citeproctext]{ref-liu2019ch3}
Liu, Jianan, Robert F. Stambaugh, and Yu Yuan. 2019. {``Size and Value
in China.''} \emph{Journal of Financial Economics} 134 (1): 48--69.
\url{https://doi.org/10.1016/j.jfineco.2019.03.008}.

\bibitem[\citeproctext]{ref-lorenzen1986}
Lorenzen, Thomas J., and Lonnie C. Vance. 1986. {``The Economic Design
of Control Charts: A Unified Approach.''} \emph{Technometrics} 28 (1):
3--10. \url{https://doi.org/10.1080/00401706.1986.10488092}.

\bibitem[\citeproctext]{ref-martinez2026gaussian}
Martinez-Taboada, Diego, and Aaditya Ramdas. 2026.
\emph{Gaussian-Efficient Testing by Betting on the Mean of Bounded
Data}. \url{https://arxiv.org/abs/2608.21694}.

\bibitem[\citeproctext]{ref-mclean2016}
McLean, R. David, and Jeffrey Pontiff. 2016. {``Does Academic Research
Destroy Stock Return Predictability?''} \emph{The Journal of Finance} 71
(1): 5--32. \url{https://doi.org/10.1111/jofi.12365}.

\bibitem[\citeproctext]{ref-newey1987}
Newey, Whitney K., and Kenneth D. West. 1987. {``A Simple, Positive
Semi-Definite, Heteroskedasticity and Autocorrelation Consistent
Covariance Matrix.''} \emph{Econometrica} 55 (3): 703--8.
\url{https://doi.org/10.2307/1913610}.

\bibitem[\citeproctext]{ref-qu2026clqt}
Qu, Bo, and Mingguang Chen. 2026. \emph{{CLQT}: A Closed-Loop,
Cost-Aware, Strategy-Consistent Benchmark for Diagnostic Evaluation of
{LLM} Portfolio-Management Agents}.
\url{https://arxiv.org/abs/2606.29771}.

\bibitem[\citeproctext]{ref-sea2026}
Sengupta, Biswa. 2026. \emph{Self-Evolving Agents with Anytime-Valid
Certificates}. \url{https://arxiv.org/abs/2607.00871}.

\bibitem[\citeproctext]{ref-pace2026}
Shawn, Zayx. 2026. \emph{{PACE}: Anytime-Valid Acceptance Tests for
Self-Evolving Agents}. \url{https://arxiv.org/abs/2606.08106}.

\bibitem[\citeproctext]{ref-shin2024edetectors}
Shin, Jaehyeok, Aaditya Ramdas, and Alessandro Rinaldo. 2024.
{``E-Detectors: A Nonparametric Framework for Sequential Change
Detection.''} \emph{The New England Journal of Statistics in Data
Science} 2 (2): 229--60. \url{https://doi.org/10.51387/23-NEJSDS51}.

\bibitem[\citeproctext]{ref-sun2026weighted}
Sun, Liulei, and Ruodu Wang. 2026. \emph{Admissibility and Complete
Classes for False Discovery Rate Control with e-Values}.
\url{https://arxiv.org/abs/2607.14380}.

\bibitem[\citeproctext]{ref-alphaagent2025}
Tang, Ziyi, Zechuan Chen, Jiarui Yang, et al. 2025. \emph{{AlphaAgent}:
{LLM}-Driven Alpha Mining with Regularized Exploration to Counteract
Alpha Decay}. \url{https://arxiv.org/abs/2502.16789}.

\bibitem[\citeproctext]{ref-ville1939}
Ville, Jean. 1939. \emph{{Étude critique de la notion de collectif}}.
Monographies Des Probabilités, Fascicule III. Gauthier-Villars.

\bibitem[\citeproctext]{ref-vovk2021merging}
Vovk, Vladimir, and Ruodu Wang. 2021. {``E-Values: Calibration,
Combination and Applications.''} \emph{The Annals of Statistics} 49 (3):
1736--54. \url{https://doi.org/10.1214/20-AOS2020}.

\bibitem[\citeproctext]{ref-wang2025stopping}
Wang, Hongjian, Sanjit Dandapanthula, and Aaditya Ramdas. 2025.
\emph{Anytime-Valid {FDR} Control with the Stopped e-{BH} Procedure}.
\url{https://arxiv.org/abs/2502.08539}.

\bibitem[\citeproctext]{ref-wang2022ebh}
Wang, Ruodu, and Aaditya Ramdas. 2022. {``False Discovery Rate Control
with e-Values.''} \emph{Journal of the Royal Statistical Society Series
B: Statistical Methodology} 84 (3): 822--52.
\url{https://doi.org/10.1111/rssb.12489}.

\bibitem[\citeproctext]{ref-waudby2024betting}
Waudby-Smith, Ian, and Aaditya Ramdas. 2024. {``Estimating Means of
Bounded Random Variables by Betting.''} \emph{Journal of the Royal
Statistical Society Series B: Statistical Methodology} 86 (1): 1--27.
\url{https://doi.org/10.1093/jrsssb/qkad009}.

\bibitem[\citeproctext]{ref-xu2026compound}
Xu, Ziyu, Lasse Fischer, and Aaditya Ramdas. 2026. \emph{Improving
Online {FDR} Procedures via Online Analogs of e-Closure and Compound
e-Values}. \url{https://arxiv.org/abs/2603.24792}.

\bibitem[\citeproctext]{ref-alphaschema2026}
Yi, Jingyang, Jian Yang, Yifei Jin, Yuqi Li, and Jian Li. 2026.
\emph{{AlphaSchema}: Exploring the Space of Trading Semantics for
{LLM}-Based Alpha Mining}. \url{https://arxiv.org/abs/2607.26642}.

\bibitem[\citeproctext]{ref-alphagen2023}
Yu, Shuo, Hongyan Xue, Xiang Ao, et al. 2023. {``Generating Synergistic
Formulaic Alpha Collections via Reinforcement Learning.''}
\emph{Proceedings of the 29th ACM SIGKDD Conference on Knowledge
Discovery and Data Mining}, 5476--86.
\url{https://doi.org/10.1145/3580305.3599831}.

\end{CSLReferences}
\end{document}